\documentclass{article} %
\usepackage{iclr2027_conference,times}

\usepackage{amsmath,amsfonts,bm}

\def\eqref#1{equation~\ref{#1}}

\def\1{\bm{1}}

\DeclareMathAlphabet{\mathsfit}{\encodingdefault}{\sfdefault}{m}{sl}
\SetMathAlphabet{\mathsfit}{bold}{\encodingdefault}{\sfdefault}{bx}{n}

\usepackage{hyperref}
\usepackage{url}
\usepackage{xspace}
\usepackage{amssymb}
\usepackage{algorithm}
\usepackage{algpseudocode}
\usepackage{graphicx}
\usepackage{booktabs}
\usepackage{multirow}
\usepackage{colortbl}
\usepackage{tabularx}
\usepackage{wrapfig}
\usepackage[most]{tcolorbox}

\definecolor{LightCyan}{rgb}{0.94, 1.0, 1.0}
\definecolor{ForestGreen}{rgb}{0.13, 0.55, 0.13}
\definecolor{LightBlue}{rgb}{0.678, 0.847, 0.902}
\definecolor{DarkRed}{rgb}{0.7, 0.1, 0.1}

\tcbset{
  aibox/.style={
    width=\linewidth,
    top=8pt,
    bottom=4pt,
    left=6pt,
    right=6pt,
    colback=blue!6!white,
    colframe=black,
    colbacktitle=black,
    fontupper=\small,
    enhanced,
    attach boxed title to top left={yshift=-0.1in,xshift=0.15in},
    boxed title style={boxrule=0pt,colframe=white},
  }
}
\newtcolorbox{AIbox}[2][]{aibox,title={#2},#1}

\title{ViSkill: Reinforcing VLM Agents with Evolving Visual-Native Skills}

\iclrfinalcopy

\author{%
  \textbf{Hongxing Li}$^{1}$, ~
  \textbf{Dingming Li}$^{1}$, ~
  \textbf{Yixin Li}$^{1}$, ~
  \textbf{Yong Du}$^{1}$, ~
  \textbf{Wenqi Zhang}$^{1}$, ~
  \textbf{Weiming Lu}$^{1}$, ~ \\
  \textbf{Jun Xiao}$^{1}$, ~ 
  \textbf{Yueting Zhuang}$^{1}$, ~
  \textbf{Yongliang Shen}$^{1}$\setcounter{footnote}{1}\thanks{Corresponding author.} \\[6pt]
  $^{1}$Zhejiang University \\ \texttt{\{hongxing.li, syl\}@zju.edu.cn}
}

\newcommand{\method}{ViSkill\xspace}

\begin{document}

\maketitle
\lhead{}
\renewcommand{\headrulewidth}{0pt}

\vspace{-10pt}
\begin{abstract}
Skill-augmented agents improve sample efficiency by distilling successful trajectories into reusable strategies. Yet most existing approaches remain text-centric, linearizing spatial layouts and action--state correspondences into language that loses critical geometric structure. Recent efforts have begun incorporating visual evidence, but construct and update skills separately from policy optimization, leaving their mutual improvement underexplored. We propose \method, a visual-native skill learning framework that encodes successful interactions as composite visual skill cards directly accessible to VLM agents. Retrieved skills guide both inference and reward shaping, while successful trajectories are distilled back into the library, forming a closed feedback loop in which skill accumulation and policy improvement reinforce each other. An optional cold-start mechanism further accelerates early-stage learning. Evaluated on Sokoban, FrozenLake, and PrimitiveSkill, \method achieves an overall success rate of $0.89$, rising to $0.91$ with cold-start initialization, outperforming all evaluated proprietary and open-source baselines while converging faster than standard PPO.
Our code is available at \texttt{https://github.com/ZJU-REAL/ViSkill}.
\end{abstract}

\section{Introduction}

\begin{wrapfigure}{r}{0.48\textwidth}
    \centering
    \includegraphics[width=\linewidth]{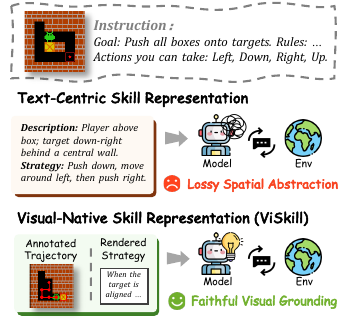}
    \vspace{-10pt}
    \caption{Text-centric skills linearize spatial experience into language, causing lossy abstraction. \method instead represents reusable experience as visual skill cards that preserve spatial structure natively.}
    \label{fig:motivation}
    \vspace{-15pt}
\end{wrapfigure}

Reinforcement learning (RL)~\citep{schulman2017ppo,guo2025deepseek-r1,ahmadian2024rloo} has become a central paradigm for training Vision-Language Model (VLM) agents that interact with complex, multi-turn environments~\citep{zhai2024fine,wang2026vagen}. Under standard RL, successful strategies are absorbed only implicitly into policy parameters, requiring the agent to rediscover similar spatial solutions across episodes~\citep{feng2026gigpo,wang2026beacon,dong2026arpo}. Yet successful trajectories already encode reusable procedural knowledge, such as how to navigate around obstacles or how to coordinate multi-step object manipulations, that could directly benefit future tasks. This motivates a natural complement to parameter-based learning: explicitly retaining and reusing successful experience across episodes.

Skill-augmented agents realize this by distilling successful trajectories into reusable strategies and retrieving them for future tasks~\citep{wang2023voyager, zhao2024expel, xia2026skillrl, shi2026skill1, zhang2026retroagent}, and have shown consistent gains in both sample efficiency and task performance. However, two fundamental gaps limit their effectiveness in visually grounded environments, where, unlike text-based settings, the agent must interpret spatial layouts, track object relationships, and reason about how actions induce visual state transitions~\citep{yang2025embodiedbench, zhai2024fine}. The first is a \textbf{representation gap}: existing skills are predominantly text-centric~\citep{ni2026trace2skill,yang2026skillopt,lin2026muse}, linearizing spatial layouts, relative positions, and action--state correspondences into language, which risks losing geometric information critical for strategy execution, as shown in Figure~\ref{fig:motivation}. The second is a \textbf{learning gap}: although recent work has begun incorporating visual evidence into skill construction and utilization~\citep{zhang2026mmskills, jiang2026xskill,liu2026skilllens,wang2026atlasva}, these approaches typically decouple skill construction from policy optimization, leaving the bidirectional interplay between visual skill acquisition and policy learning largely underexplored.

We introduce \textbf{\method}, a visual-native skill learning framework that couples reusable visual experience with reinforcement optimization for VLM agents. Each skill is stored as a composite visual skill card that renders annotated trajectory frames alongside a policy-distilled strategy description into a unified image, directly preserving the spatial and procedural context of successful experience. Given the current task state, \method retrieves skills by matching geometric configurations and injects the full skill card into the VLM's visual context. The retrieved skill further provides a skill-guided reward that assigns partial credit to failed trajectories that show skill-relevant progress. After each episode, successful trajectories that pass quality and novelty gates are distilled back into the library as new visual skills. To bootstrap this loop, a cold-start mechanism optionally pre-populates the library with a small set of diverse, solver-derived seed skills. Together, these mechanisms form a closed feedback loop: retrieved skills improve the policy through in-context demonstration and reward shaping, while the improved policy contributes higher-quality skills back to the library.

We evaluate \method on Sokoban, FrozenLake, and PrimitiveSkill, spanning discrete 2D spatial planning and coordinate-grounded 3D manipulation. \method outperforms all evaluated proprietary and open-source baselines with an overall success rate of $0.89$, rising to $0.91$ with cold-start initialization. Beyond final performance, \method converges faster and achieves higher success rates throughout training, reflecting the positive feedback loop between skill accumulation and policy improvement. Ablations further confirm that visual skill representations consistently outperform text-centric alternatives, and that skills distilled under joint policy optimization transfer more effectively than those accumulated without it.

Overall, our contributions are threefold:

\begin{itemize}
    \item We propose \method, a visual-native skill learning framework that represents reusable procedural experience as composite visual skill cards, preserving spatial and procedural context in a form directly accessible to VLM agents.
    \item We develop geometry-aware skill retrieval and a skill-guided reinforcement objective that form a closed feedback loop, where retrieved skills guide policy optimization and successful trajectories are distilled back into the library.
    \item We empirically demonstrate across 2D spatial planning and 3D manipulation that \method outperforms proprietary and open-source baselines, converging faster and achieving higher final success rates than standard PPO.
\end{itemize}

\section{Related Work}

\subsection{Reinforcement Learning for VLM Agents}
Recent work has extended reinforcement learning from language-model alignment to vision-language reasoning, using verifiable rewards to improve visual perception and reasoning~\citep{liu2025visual-rft,shen2025vlm-r1,li2026p2r,huang2026vision-r1,li2026spatialladder}. Beyond single-turn prediction, several studies formulate VLMs as sequential decision-making agents and optimize them through interaction with visual environments~\citep{zhai2024fine,liu2025visual-arft,zheng2026deepeyes,wang2026vagen}, addressing challenges such as long-horizon credit assignment and sparse reward in visually grounded tasks~\citep{wang2025ragen,feng2026gigpo,wang2026beacon}. However, interaction experience is typically consumed as transient rollout data and absorbed implicitly into policy parameters. Rather than treating experience solely as a training signal, \method externalizes successful trajectories into an explicit library of retrievable visual-native skills, allowing accumulated experience to guide future inference and provide dense rewards that facilitate policy optimization.

\subsection{Skill-Augmented Agents}
Skills provide a modular mechanism for transferring procedural knowledge across episodes. Most existing approaches distill interaction trajectories into natural-language instructions or structured workflows organized into persistent libraries for retrieval and reuse~\citep{shinn2023reflexion,zhao2024expel,wang2023voyager}, with recent work further incorporating such libraries into reinforcement learning~\citep{xia2026skillrl,shi2026skill1,lu2026skill0}. However, these approaches remain text-centric, linearizing spatial layouts and action--state correspondences that are central to visually grounded decision making. Recent methods have begun incorporating visual information into skill construction and memory~\citep{jiang2026xskill,zhang2026mmskills,huang2026skill-cmib,liu2026skilllens}. AtlasVA~\citep{wang2026atlasva} further evolves spatial atlases from interaction statistics for dense reward shaping. Within this broader direction, \method represents reusable procedures as composite visual skill cards and couples their online distillation with policy optimization.

\section{Methodology}

\begin{figure}
    \centering
    \includegraphics[width=1\linewidth]{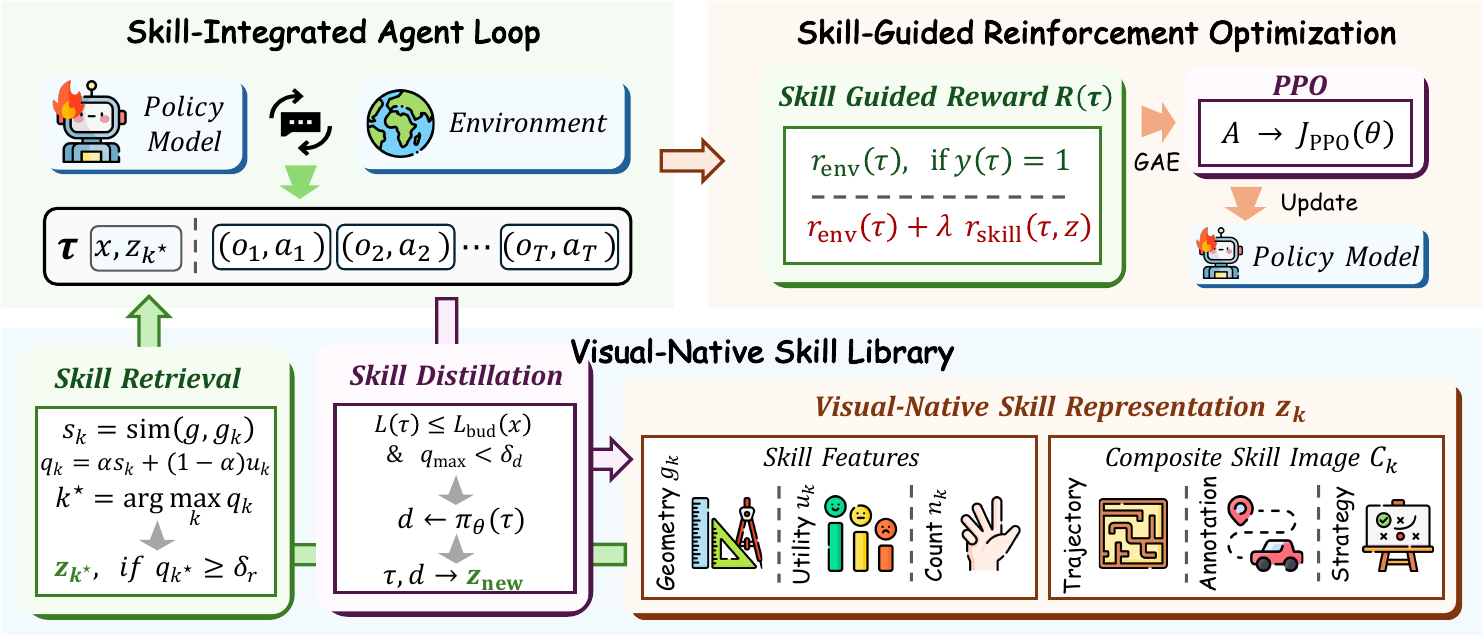}
    \vspace{-10pt}
    \caption{Overview of \method. The agent retrieves a visual skill $z_{k^\star}$ from the skill library to guide interaction, collects trajectory $\tau$, and distills eligible trajectories into new skills. The skill-guided reward $R(\tau)$ and PPO jointly optimize the policy.}
    \label{fig:method}
    \vspace{-10pt}
\end{figure}

\subsection{Skill-Integrated Agent Loop}
Figure~\ref{fig:method} provides an overview of \method. For each task $x \sim \mathcal{D}$, \method proceeds through three stages: skill retrieval, skill-guided interaction, and online skill distillation. A complete interaction trajectory is represented as $\tau=(x,z,(o_t,a_t)_{t=1}^{T})$, where $z$ denotes the retrieved visual skill and $(o_t,a_t)_{t=1}^{T}$ denotes the sequence of observations and actions. During interaction, the policy $\pi_\theta$ generates actions conditioned on the task, interaction history, and the retrieved skill. Upon termination, the environment provides a binary success indicator $y(\tau)\in\{0,1\}$, which serves as the outcome reward for policy optimization, updates the utility of the retrieved skill, and determines whether the trajectory is eligible for distillation into a new skill.

\subsection{Visual-Native Skill Library}

\paragraph{Visual-Native Skill Representation.}
Textual skill descriptions often lose the spatial and procedural structure needed for visually grounded decision making. To preserve this information in a form directly accessible to VLMs, we maintain a visual skill library $\mathcal{B}=\{z_k\}_{k=1}^{|\mathcal{B}|}$, where each skill $z_k$ is constructed from a successful interaction trajectory $\hat{\tau}_k$ through a structured rendering process. We retain the original visual observations, augment them with trajectory annotations to clarify spatial and procedural progression, and organize them into a unified visual sequence. A strategy description $d_k$, distilled by the policy from $\hat{\tau}_k$, is rendered as text within the same image to capture the high-level intent of the skill. The annotated trajectory and rendered strategy are then composed into a single composite image $C_k$, which is directly provided to the VLM as visual context. Formally:
\begin{equation}
    z_k=\bigl(C_k,g_k,u_k,n_k\bigr),
    \qquad
    C_k=\operatorname{Render}\!\left(
        d_k,\operatorname{Annotate}(\hat{\tau}_k)
    \right),
\end{equation}
where $C_k$ denotes the composite skill card, $g_k$ is the geometric descriptor used for skill retrieval, and $u_k$ and $n_k$ are its utility and usage count, respectively.

\paragraph{Skill Retrieval and Utilization.}
Text-centric skill libraries often retrieve skills through language-level matching between the current instruction and textual skill descriptions~\citep{shi2026skill1,xia2026skillrl,zhang2025qwen3embedding}. In visually grounded tasks, however, similar instructions may correspond to different spatial layouts that require distinct strategies. \method therefore determines skill relevance through geometric similarity rather than language matching, while incorporating historical utility to account for skill reliability. At the beginning of each episode, an environment-specific feature extractor maps the initial task configuration to a geometric descriptor $g$, capturing spatial relations such as relative positions, distances, and obstacles. For each skill $z_k$, we compute a composite score:
\begin{equation}
    q_k=\alpha s_k+(1-\alpha)u_k,
\end{equation}
where $s_k=\operatorname{sim}(g,g_k)$ is the geometric similarity and $u_k$ is the historical utility. We first identify the highest-scoring skill as $k^\star=\arg\max_k q_k$. The corresponding skill $z_{k^\star}$ is retrieved only if $q_{k^\star}\geq\delta_r$; otherwise, the policy proceeds without skill guidance. The retrieved composite card $C_{k^\star}$ is directly appended to the VLM's visual context. After termination, the utility and usage count are updated as follows:
\begin{equation}
    u_{k^\star}\leftarrow(1-\beta)u_{k^\star}+\beta\, y(\tau),
    \qquad
    n_{k^\star}\leftarrow n_{k^\star}+1,
\end{equation}
where $\beta$ is the EMA coefficient and $y(\tau)$ is the binary success indicator.

\paragraph{Skill Distillation and Library Update.}
Indiscriminately storing every successful trajectory risks introducing suboptimal and redundant skills, degrading both the quality and diversity of the library. After each episode, a successful trajectory is therefore considered for distillation only if it passes both quality and novelty gates:
\begin{equation}
    L(\tau) \leq L_{\mathrm{bud}}(x), \qquad q_{\max} < \delta_d,
\end{equation}
where $L_{\mathrm{bud}}(x)$ is the trajectory-length budget for task $x$, and $q_{\max}=\max_{z_k\in\mathcal{B}}q_k$ is the best-match score used as a proxy for library coverage. The quality gate ensures that only high-quality trajectories are distilled into the library; we instantiate $L_{\mathrm{bud}}(x)=L^\star(x)$ using a task-specific solver. The novelty gate admits a trajectory only when no existing skill sufficiently covers the current configuration; for an empty library we define $q_{\max}=0$. 

For each eligible trajectory, the policy produces a reusable high-level strategy description by summarizing its own successful interaction, which is then rendered together with the annotated trajectory to construct a new skill $z_{\mathrm{new}}$. The new skill is added to the library with an initial utility and usage count. When the library reaches its capacity, the skill with the lowest utility--frequency score is evicted:
\begin{equation}
    k_{\mathrm{evict}} = \arg\min_k\, u_k \log(n_k + 1).
\end{equation}

\subsection{Skill-Guided Reinforcement Optimization}

\paragraph{Cold-Start Initialization.}
When starting from an empty library, early-stage exploration lacks skill guidance, delaying the positive feedback loop between skill distillation and policy learning. To address this cold-start problem, \method optionally pre-populates the library with a diverse set of seed skills before training begins. Candidate trajectories are generated using offline task-specific solvers with strategy descriptions distilled by an external model, and selected through farthest-point sampling in the geometric feature space to maximize initial coverage. These seed skills serve only as an initialization: once training begins, all subsequent skills are distilled entirely by the policy from its own successful trajectories. The cold-start library size is set to $N_{\mathrm{cold}}=\lfloor\gamma N_{\mathrm{conv}}\rfloor$, where $\gamma\in[0,1]$ is the cold-start ratio and $N_{\mathrm{conv}}$ is the converged library size from a preliminary run without cold-start initialization; setting $\gamma=0$ disables external-model initialization, providing a flexible trade-off between early-stage sample efficiency and independence from external-model supervision.

\algrenewcommand{\algorithmiccomment}[1]{\hfill\textcolor{ForestGreen}{$\triangleright$ \textit{#1}}}
\begin{algorithm}[H]
\caption{Training Procedure of \method}
\label{alg:viskill}
\small
\begin{algorithmic}[1]
\Require Policy $\pi_\theta$, skill library $\mathcal{B}$, task distribution $\mathcal{D}$, library capacities $N_{\mathrm{cold}},N_{\max}$, thresholds $\delta_r,\delta_d$
\If{cold-start initialization is enabled}
    \State $\mathcal{B}\gets\Call{ColdStart}{N_{\mathrm{cold}}}$ \Comment{Cold-start initialization}
\EndIf
\For{batch of $M$ tasks $\{x_i\}_{i=1}^{M}\sim\mathcal{D}$}
    \For{sample $i=1,\ldots,M$} \Comment{Skill-integrated agent loop}
        \State $g_i\gets\Call{FeatureExtractor}{x_i}$; $(z_i,q_{\max})\gets\Call{Retrieve}{g_i,\mathcal{B},\delta_r}$
        \State $\tau_i\sim\pi_\theta(\cdot\mid x_i,z_i)$ \Comment{Skill retrieval and utilization}
        \State $R(\tau_i)\gets\Call{SkillGuidedReward}{\tau_i,z_i}$ \Comment{Skill-guided reward}
        \State $\mathcal{B}\gets\Call{UpdateUtility}{\mathcal{B},z_i,y(\tau_i)}$
        \If{$y(\tau_i)=1$ and $\Call{PassGates}{\tau_i,x_i,q_{\max},\delta_d}$}
            \State $z_{\mathrm{new},i}\gets\Call{DistillVisualSkill}{\pi_\theta,\tau_i,g_i}$
            \State $\mathcal{B}\gets\Call{UpdateLibrary}{\mathcal{B},z_{\mathrm{new},i},N_{\max}}$ \Comment{Skill distillation and library update}
        \EndIf
    \EndFor
    \State $\theta\gets\Call{PPOUpdate}{\theta,\{\tau_i,R(\tau_i)\}_{i=1}^{M}}$ \Comment{Policy optimization}
\EndFor
\end{algorithmic}
\end{algorithm}
\vspace{-13pt}

\paragraph{Skill-Guided Reward.}
Task-success rewards provide little learning signal for failed trajectories, even when the agent partially follows a useful retrieved skill. We therefore introduce an auxiliary reward based on the agreement between the agent trajectory and the retrieved skill. Let $\ell$ denote the furthest point along the reference trajectory that the agent trajectory visits, and let $L(\tau)$ and $L_z$ denote the lengths of the executed and reference trajectories. The trajectory agreement is defined as:

\begin{equation}
    r_{\mathrm{skill}}(\tau,z)
    =
    \max\left(
        0,\,
        \frac{\ell-\left(L(\tau)-\ell\right)}{L_z}
    \right).
\end{equation}
The final reward combines the environment reward $r_{\mathrm{env}}(\tau)$, which includes formatting and task-completion signals, with the skill-guidance term:
\begin{equation}
    R(\tau)
    =
    \begin{cases}
        r_{\mathrm{env}}(\tau)+\lambda\,r_{\mathrm{skill}}(\tau,z),
        & y(\tau)=0 \text{ and } z \text{ is retrieved},\\
        r_{\mathrm{env}}(\tau),
        & \text{otherwise},
    \end{cases}
\end{equation}
where $\lambda$ controls the strength of skill guidance. Let $r_{\mathrm{succ}}>0$ denote the terminal bonus for task completion. Since $r_{\mathrm{skill}}\in[0,1]$ and $\lambda<r_{\mathrm{succ}}$, the skill-guidance reward is always smaller than the completion bonus, providing bounded partial credit for skill-aligned failed trajectories without distorting the global reward ordering.

\vspace{-3pt}
\paragraph{Policy Optimization.}
Given the resulting trajectory rewards, we optimize the VLM policy $\pi_\theta$ using Proximal Policy Optimization (PPO)~\citep{schulman2017ppo}. Denote by $\bar{\tau}$ the tokenized representation of trajectory $\tau$ with $N$ tokens, and let $\rho_i(\theta)=\pi_\theta(\bar{\tau}_i\mid\bar{\tau}_{<i})/\pi_{\mathrm{old}}(\bar{\tau}_i\mid\bar{\tau}_{<i})$ be the probability ratio between the current and rollout policies. The clipped PPO objective is:
\begin{equation}
    J_{\mathrm{PPO}}(\theta)
    =
    \frac{1}{\sum_{i=1}^{N}m_i}
    \sum_{i=1}^{N}m_i
    \min\!\left(
        \rho_i(\theta)A_i,
        \operatorname{clip}\!\left(
            \rho_i(\theta),1-\epsilon,1+\epsilon
        \right)A_i
    \right),
\end{equation}
where $m_i$ masks policy-generated tokens from observation tokens, $A_i$ is the advantage estimated using generalized advantage estimation, and $\epsilon$ is the clipping coefficient. Optimizing this objective drives policy improvement while the skill library grows concurrently through distillation, enabling the two components to co-evolve as interaction experience accumulates.

The complete training procedure is summarized in Algorithm~\ref{alg:viskill}.

\vspace{-3pt}
\section{Experiments}

\subsection{Experimental Setup}

\paragraph{Environments.}
We evaluate \method on three visually grounded interactive environments spanning 2D spatial reasoning and 3D embodied manipulation, adopted from VAGEN~\citep{wang2026vagen}. Sokoban requires the agent to push all boxes onto target locations; FrozenLake requires it to navigate to a goal while avoiding holes. Both environments provide grid-based visual observations and discrete directional actions. PrimitiveSkill evaluates coordinate-grounded manipulation in third-person 3D scenes, comprising the Place, Stack, Drawer, and Align tasks from VAGEN and the Swap task introduced by AtlasVA~\citep{wang2026atlasva}. We report success rate across all environments.

\vspace{-4pt}
\paragraph{Baselines.}
We compare \method against representative proprietary and open-source VLMs. Proprietary baselines include frontier models from OpenAI, Google, and Anthropic~\citep{openai2025gpt5,openai2025o3o4mini,hurst2024gpt,google2025gemini25,google2024gemini20,anthropic2025claude45sonnet,anthropic2025claude37sonnet}. For open-source models, we evaluate the Qwen2.5-VL family at 3B, 7B, and 72B scales~\citep{bai2025qwen25vltechnicalreport}, along with three reinforcement-learning-enhanced 3B baselines. VLM-R1 applies R1-style reinforcement learning with rule-based rewards to enhance general visual reasoning~\citep{shen2025vlm-r1}. VAGEN trains multi-turn VLM agents by explicitly reinforcing world-model reasoning~\citep{wang2026vagen}. AtlasVA~\citep{wang2026atlasva} is a memory-augmented RL baseline that evolves danger and affinity heatmaps from interaction statistics and uses them for dense spatial reward shaping. For a controlled comparison, VAGEN, AtlasVA, and \method share the same Qwen2.5-VL-3B backbone and environment seeds. \method follows the training horizon of VAGEN, which is shorter than that of AtlasVA.

\vspace{-4pt}
\paragraph{Implementation Details.}
We use Qwen2.5-VL-3B-Instruct as the policy backbone, following the PPO configuration of VAGEN~\citep{wang2026vagen} with a rollout batch size of 128, a PPO mini-batch size of 32, and actor and critic learning rates of $1\times10^{-6}$ and $1\times10^{-5}$. Sokoban and FrozenLake are trained for 300 steps on $4\times$ H100 GPUs; PrimitiveSkill is trained jointly for 60 steps on $2\times$ H100 GPUs with an additional H100 GPU dedicated to the ManiSkill~\citep{mu2021maniskill} environment server. We set the similarity weight $\alpha=0.9$, EMA coefficient $\beta=0.1$, skill-guidance weight $\lambda=0.8$, novelty threshold $\delta_d=0.9$, retrieval threshold $\delta_r=0.8$, and cold-start ratio $\gamma=0.2$, with GPT-5.4-mini~\citep{openai2026gpt54mini} serving as the external model for cold-start strategy distillation.

\subsection{Main Results}

\begin{table*}[t!]
\centering
\small
\resizebox{\textwidth}{!}{
\begin{tabular}{lccccccccc}
\toprule
\multirow{2}{*}{\textbf{Method}}
& \multirow{2}{*}{\textbf{Sokoban}}
& \multirow{2}{*}{\textbf{FrozenLake}}
& \multicolumn{6}{c}{\textbf{PrimitiveSkill}}
& \multirow{2}{*}{\textbf{Overall}} \\
\cmidrule(lr){4-9}
& & & \textbf{Place} & \textbf{Stack} & \textbf{Drawer}
& \textbf{Align} & \textbf{Swap} & \textbf{Average} & \\
\midrule
\rowcolor{gray!10}
\multicolumn{10}{l}{\textit{\textbf{Proprietary Models}}} \\
GPT-5                         & 0.70 & 0.77 & 1.00 & 0.63 & 0.00 & 1.00 & 0.55 & 0.64 & 0.70 \\
o3                            & 0.60 & 0.78 & 1.00 & 0.63 & 0.00 & 1.00 & 0.82 & 0.69 & 0.69 \\
o4-mini                       & 0.44 & 0.82 & 1.00 & 0.50 & 0.00 & 0.75 & 0.33 & 0.52 & 0.59 \\
GPT-4o                        & 0.43 & 0.54 & 0.50 & 0.63 & 0.00 & 0.88 & 0.94 & 0.59 & 0.52 \\
Gemini 2.5 Pro                & 0.58 & 0.78 & 0.63 & 0.63 & 0.00 & 0.75 & 0.00 & 0.40 & 0.59 \\
Gemini 2.0                    & 0.28 & 0.61 & 0.75 & 0.13 & 0.00 & 0.25 & 0.32 & 0.29 & 0.39 \\
Claude 4.5 Sonnet             & 0.31 & 0.80 & 0.63 & 0.50 & 0.00 & 1.00 & 1.00 & 0.63 & 0.58 \\
Claude 3.7 Sonnet             & 0.25 & 0.69 & 0.63 & 0.13 & 0.00 & 1.00 & 0.90 & 0.53 & 0.49 \\
\midrule
\rowcolor{gray!10}
\multicolumn{10}{l}{\textit{\textbf{Open-Source Models}}} \\
Qwen2.5-VL-72B & 0.18 & 0.44 & 1.00 & 0.50 & 0.00 & 1.00 & 0.33 & 0.57 & 0.40 \\
Qwen2.5-VL-7B  & 0.13 & 0.14 & 0.00 & 0.00 & 0.00 & 0.75 & 0.03 & 0.16 & 0.14 \\
Qwen2.5-VL-3B  & 0.14 & 0.14 & 0.00 & 0.00 & 0.00 & 0.00 & 0.00 & 0.00 & 0.09 \\
VLM-R1-3B      & 0.13 & 0.13 & 0.00 & 0.00 & 0.00 & 0.00 & 0.00 & 0.00 & 0.09 \\
VAGEN          & 0.79 & 0.74 & 1.00 & 0.88 & 1.00 & 1.00 & 0.50 & \underline{0.88} & 0.80 \\
Atlas-VA       & 0.79 & 0.83 & 1.00 & 1.00 & 1.00 & 1.00 & 1.00 & \textbf{1.00} & 0.87 \\
\midrule
\rowcolor{gray!10}
\multicolumn{10}{l}{\textit{\textbf{\method (Ours)}}} \\
\method              & \underline{0.82} & \underline{0.84} & 1.00 & 1.00 & 1.00 & 1.00 & 1.00 & \textbf{1.00} & \underline{0.89} \\
\method + \textit{Cold-Start} & \textbf{0.88} & \textbf{0.85} & 1.00 & 1.00 & 1.00 & 1.00 & 1.00 & \textbf{1.00} & \textbf{0.91} \\
\bottomrule
\end{tabular}
}
\vspace{-5pt}
\caption{Success rates on Sokoban, FrozenLake, and PrimitiveSkill. \textbf{Bold} and \underline{underlined} entries indicate the best and second-best results, respectively.}
\vspace{-10pt}
\label{tab:main_results}
\end{table*}

\method outperforms all proprietary and open-source baselines in overall success rate, achieving $0.89$ (Table~\ref{tab:main_results}). On the 2D environments, it obtains $0.82$ on Sokoban and $0.84$ on FrozenLake, surpassing both VAGEN ($0.79$, $0.74$), which relies on world-model reasoning, and AtlasVA ($0.79$, $0.83$), which employs spatially grounded reward shaping. On PrimitiveSkill, \method achieves a perfect average success rate of $1.00$, matching AtlasVA and outperforming VAGEN by $12$ percentage points. These results demonstrate that explicitly retaining and reusing successful visual experience provides an effective learning signal while also offering actionable guidance at both training and inference time, enabling the policy to ground its decisions in previously successful strategies across both discrete 2D planning and coordinate-grounded 3D manipulation.

Cold-start initialization further raises the overall success rate from $0.89$ to $0.91$, exceeding AtlasVA ($0.87$) by $4$ percentage points. The gains are concentrated in the 2D environments: performance improves from $0.82$ to $0.88$ on Sokoban and from $0.84$ to $0.85$ on FrozenLake, while PrimitiveSkill remains saturated at $1.00$. The larger improvement on Sokoban suggests that diverse seed skills are particularly beneficial for combinatorial configurations, reducing unguided early-stage exploration and providing broader initial coverage for subsequent policy learning.

\subsection{Training Dynamics}

\begin{figure*}[t]
    \centering
    \includegraphics[width=\textwidth]{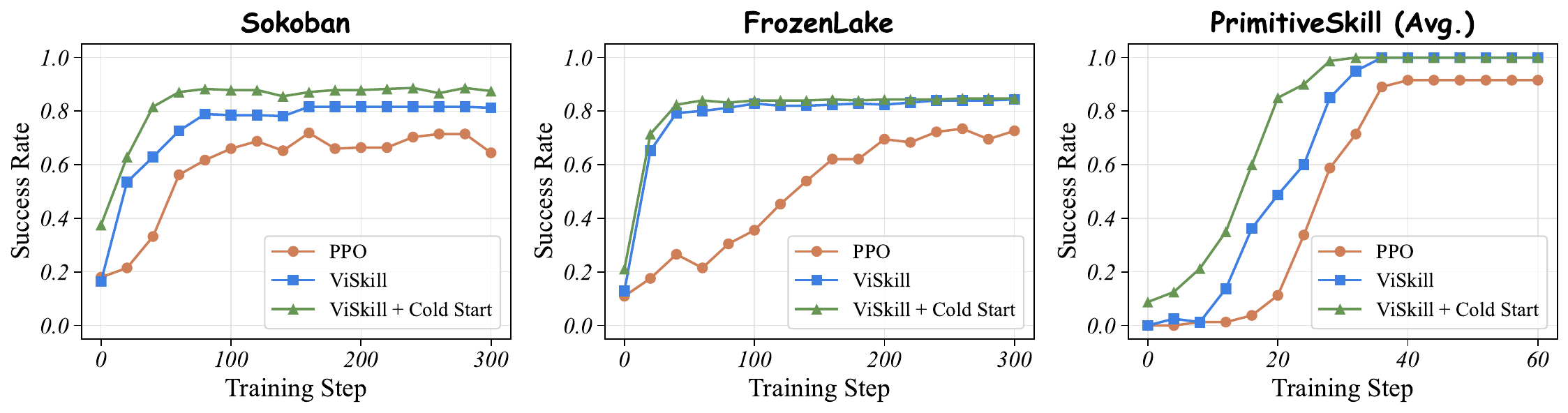}
    \vspace{-13pt}
    \caption{Training dynamics of PPO, \method, and \method + \textit{Cold-Start} across Sokoban, FrozenLake, and PrimitiveSkill. \method converges faster and achieves higher success rates than PPO across all environments, with cold-start initialization providing additional gains in early-stage learning.}
    \label{fig:training_dynamics}
\end{figure*}

Figure~\ref{fig:training_dynamics} presents the training dynamics of PPO, \method, and \method + \textit{Cold-Start} across all three environments. \method consistently converges faster and reaches higher success rates than PPO, demonstrating the benefit of retaining and reusing successful visual experience as an explicit training signal. Cold-start initialization further accelerates early-stage learning: on Sokoban, \method + \textit{Cold-Start} achieves substantially higher success rates from the first few steps and maintains its lead throughout training, suggesting that diverse seed skills provide broader initial coverage for combinatorial configurations. On PrimitiveSkill, both \method variants converge to 1.00 significantly faster than PPO, with cold-start initialization providing a notable advantage in the earliest training steps.

\begin{table*}[t]
\centering
\begin{minipage}[t]{0.4\textwidth}
    \centering
    \small
    \setlength{\tabcolsep}{4pt}
    \begin{tabularx}{\linewidth}{@{}Xcc@{}}
        \toprule
        \textbf{Method} & \textbf{Sokoban} & \textbf{FrozenLake} \\
        \midrule
        \rowcolor{gray!10}
        \method   & \textbf{0.82} & \textbf{0.84} \\
        Text-Skill & $0.63_{\textcolor{DarkRed}{\text{\textit{-0.19}}}}$ & $0.59_{\textcolor{DarkRed}{\text{\textit{-0.25}}}}$ \\
        OCR-Skill  & $0.68_{\textcolor{DarkRed}{\text{\textit{-0.14}}}}$ & $0.62_{\textcolor{DarkRed}{\text{\textit{-0.22}}}}$ \\
        \bottomrule
    \end{tabularx}
    \caption{Comparison of skill representations on Sokoban and FrozenLake.}
    \vspace{-5pt}
    \label{tab:skill_representation}
\end{minipage}
\hfill
\begin{minipage}[t]{0.56\textwidth}
    \centering
    \small
    \setlength{\tabcolsep}{4pt}
    \begin{tabularx}{\linewidth}{@{}Xcc@{}}
        \toprule
        \textbf{Method} & \textbf{Sokoban} & \textbf{FrozenLake} \\
        \midrule
        \method \textit{w/o Skill-Guided Reward}    & $0.77_{\textcolor{DarkRed}{\text{\textit{-0.05}}}}$ & $0.80_{\textcolor{DarkRed}{\text{\textit{-0.04}}}}$ \\
        \method \textit{w/o Strategy}     & $0.72_{\textcolor{DarkRed}{\text{\textit{-0.10}}}}$ & $0.81_{\textcolor{DarkRed}{\text{\textit{-0.03}}}}$ \\
        \method \textit{w/o Optimization} & $0.29_{\textcolor{DarkRed}{\text{\textit{-0.53}}}}$ & $0.33_{\textcolor{DarkRed}{\text{\textit{-0.51}}}}$ \\
        \bottomrule
    \end{tabularx}
    \caption{Ablation of \method components on Sokoban and FrozenLake.}
    \vspace{-5pt}
    \label{tab:component_ablation}
\end{minipage}
\end{table*}

\begin{figure*}[t]
    \centering
    \begin{minipage}[t]{0.49\textwidth}
        \centering
        \includegraphics[width=\linewidth]{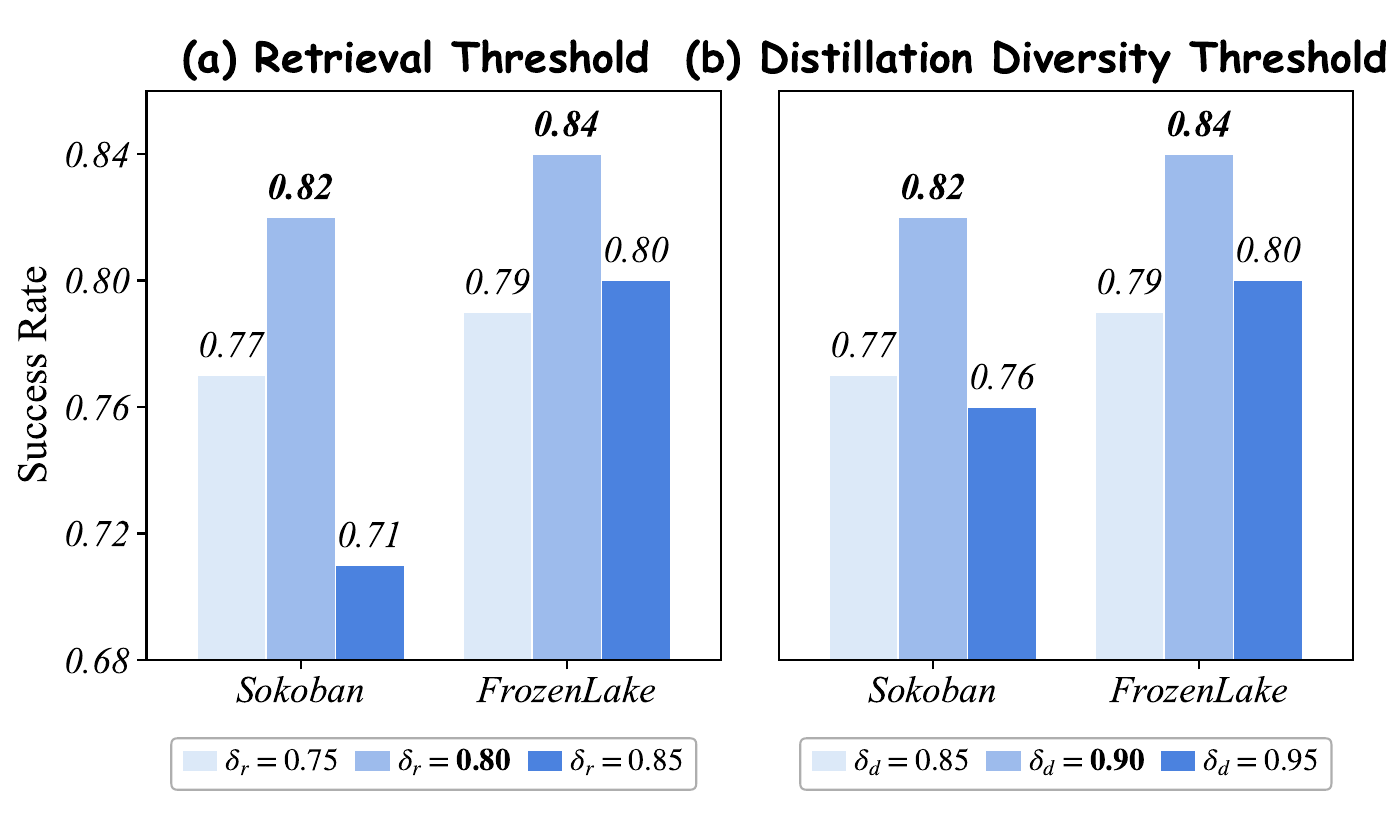}
        \vspace{-5pt}
        \caption{Impact of the retrieval threshold $\delta_r$ and distillation diversity threshold $\delta_d$ on success rate in Sokoban and FrozenLake.}
        \vspace{-6pt}
        \label{fig:threshold_ablation}
    \end{minipage}
    \hfill
    \begin{minipage}[t]{0.49\textwidth}
        \centering
        \includegraphics[width=\linewidth]{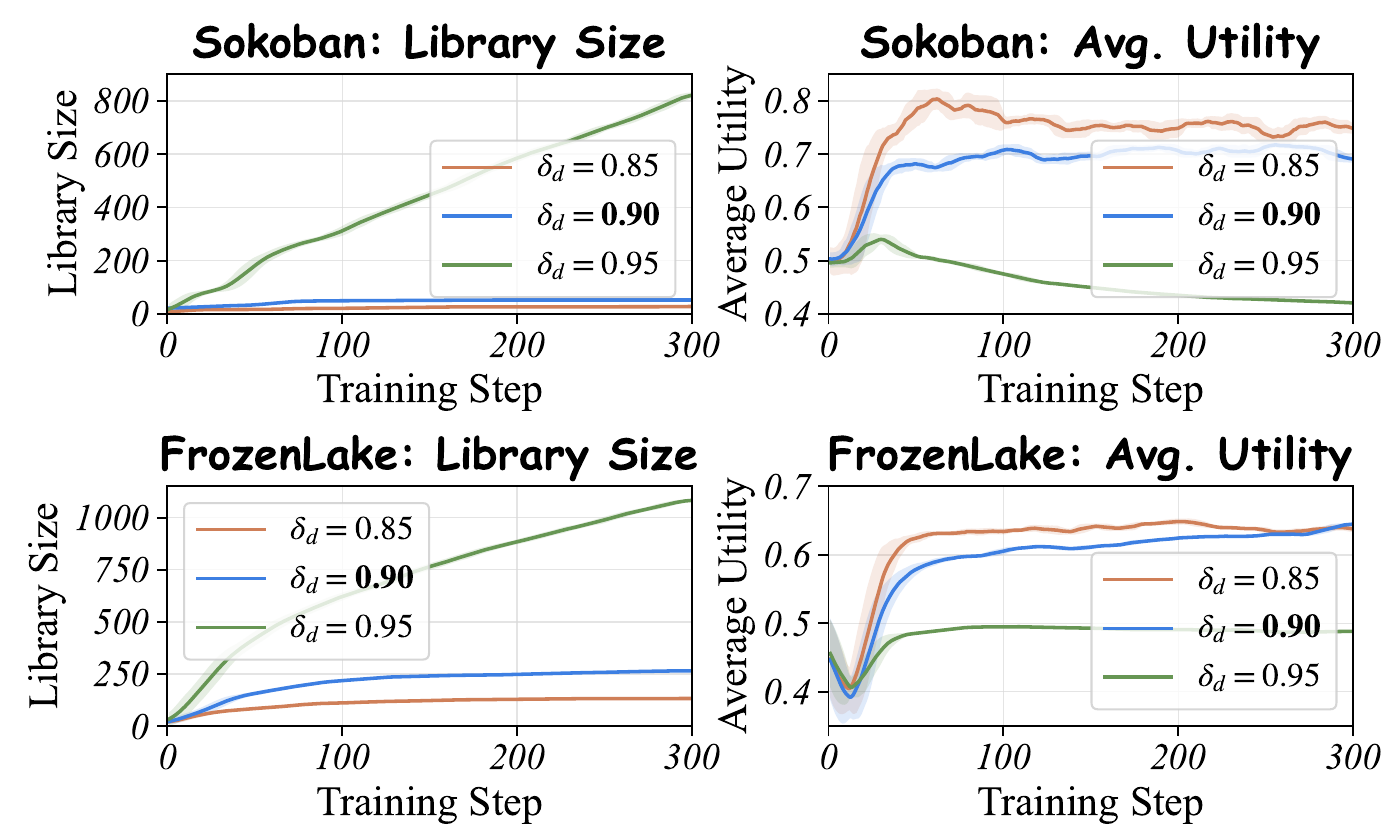}
        \vspace{-5pt}
        \caption{Skill-library dynamics under different distillation diversity thresholds $\delta_d$, measured by library size and average utility.}
        \vspace{-6pt}
        \label{fig:skill_library_dynamics_2x2}
    \end{minipage}
\end{figure*}

\subsection{Ablation Study}

\paragraph{Skill Representation.}
Since PrimitiveSkill is saturated by the full model, we conduct controlled ablations on Sokoban and FrozenLake. Table~\ref{tab:skill_representation} compares three skill representation strategies. Text-Skill and OCR-Skill, which encode successful strategies as raw text and OCR-rendered images respectively, both underperform \method by a substantial margin across Sokoban and FrozenLake
. This suggests that preserving the original visual trajectory as the skill representation is essential, as it retains the spatial and structural context of successful strategies that text-based representations inevitably lose, even when rendered back into image form.

\paragraph{\method Component.}
Table~\ref{tab:component_ablation} ablates the three core components of \method. Removing Skill-Guided Reward leads to moderate performance drops ($-0.05$ on Sokoban, $-0.04$ on FrozenLake), indicating that the skill-derived reward signal provides a useful but not dominant training contribution. Removing Strategy causes a larger drop on Sokoban ($-0.10$), suggesting that high-level strategic guidance is particularly beneficial for combinatorial planning tasks. Removing Optimization yields largest degradation ($-0.53$ on Sokoban, $-0.51$ on FrozenLake), where skill retrieval and distillation are retained but model training is disabled. This highlights the importance of the co-evolution between policy optimization and skill learning: training enables the model to better leverage retrieved skills, while an improving policy in turn contributes higher-quality skills to the library, creating a mutually reinforcing cycle that is essential for effective learning.

\paragraph{Skill-Guided Reward Weight.}
\begin{wraptable}{r}{0.45\columnwidth}
    \centering
    \footnotesize
    \begin{tabular*}{0.9\linewidth}{@{\extracolsep{\fill}}lcccc@{}}
        \toprule
        $\boldsymbol{\lambda}$ & $\mathbf{0.0}$ & $\mathbf{0.6}$ & $\mathbf{0.8}$ & $\mathbf{1.0}$ \\
        \midrule
        Sokoban    & $0.77$ & $0.71$ & \textbf{0.82} & $0.77$ \\
        FrozenLake & $0.80$ & $0.79$ & \textbf{0.84} & $0.81$ \\
        \bottomrule
    \end{tabular*}
    \caption{Impact of $\lambda$ on success rate.}
    \label{tab:guided_reward_weight}
\end{wraptable}

Table~\ref{tab:guided_reward_weight} studies the sensitivity to the skill-guided reward weight $\lambda$. Performance peaks at $\lambda=0.8$ on both environments, achieving $0.82$ on Sokoban and $0.84$ on FrozenLake. Setting $\lambda=0.0$ removes the skill-guided signal entirely and leads to a drop of $0.05$ on Sokoban, while setting $\lambda=1.0$ over-emphasizes progress toward retrieved reference states and similarly degrades performance. An intermediate value of $\lambda=0.8$ strikes the best balance, effectively complementing the environment reward with retrieved skill guidance without suppressing policy exploration. 

\paragraph{Retrieval Threshold.}
Figure~\ref{fig:threshold_ablation} (a) shows the effect of the retrieval threshold $\delta_r$ on success rate. Performance peaks at $\delta_r=0.80$ on both Sokoban ($0.82$) and FrozenLake ($0.84$). A lower threshold ($\delta_r=0.75$) admits skills whose geometric states are poorly matched to the current observation, introducing misaligned guidance that degrades performance. A higher threshold ($\delta_r=0.85$) is overly restrictive, reducing the frequency of successful retrievals and limiting the influence of skill guidance during training. These results confirm that $\delta_r=0.80$ strikes the right balance between retrieval relevance and coverage.

\paragraph{Distillation Diversity Threshold.}
Figure~\ref{fig:threshold_ablation} (b) and Figure~\ref{fig:skill_library_dynamics_2x2} examine the effect of the novelty threshold $\delta_d$, which gates library admission based on the similarity between a new trajectory and existing skills. A low threshold ($\delta_d=0.85$) is overly permissive: the library grows rapidly throughout training, but the influx of redundant skills causes average utility to decline steadily, ultimately hurting success rate. A high threshold ($\delta_d=0.95$) enforces strict diversity, keeping utility high but severely limiting library growth and reducing the range of scenarios the library can cover. The intermediate setting $\delta_d=0.90$ maintains a moderate library size with consistently high utility, achieving the best success rate on both environments.

\subsection{Further Analysis}

\begin{figure*}[t]
    \centering
    \begin{minipage}[t]{0.34\textwidth}
        \centering
        \includegraphics[width=\linewidth]{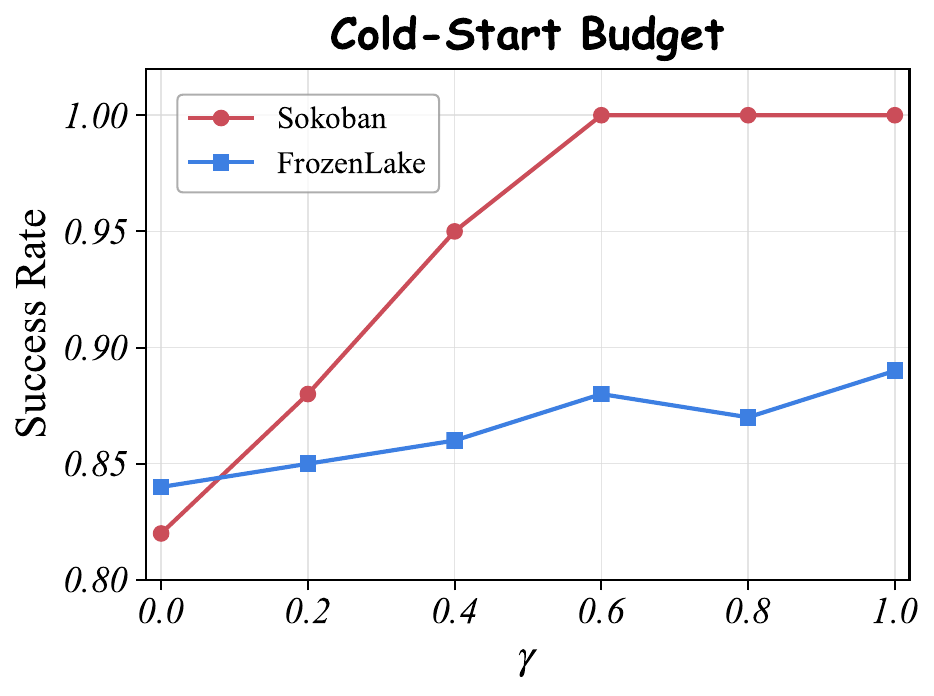}
        \vspace{-10pt}
        \caption{Impact of the cold-start ratio $\gamma$ on success rate in Sokoban and FrozenLake.}
        \label{fig:cold_start_budget}
    \end{minipage}\hfill
    \begin{minipage}[t]{0.64\textwidth}
        \centering
        \includegraphics[width=\linewidth]{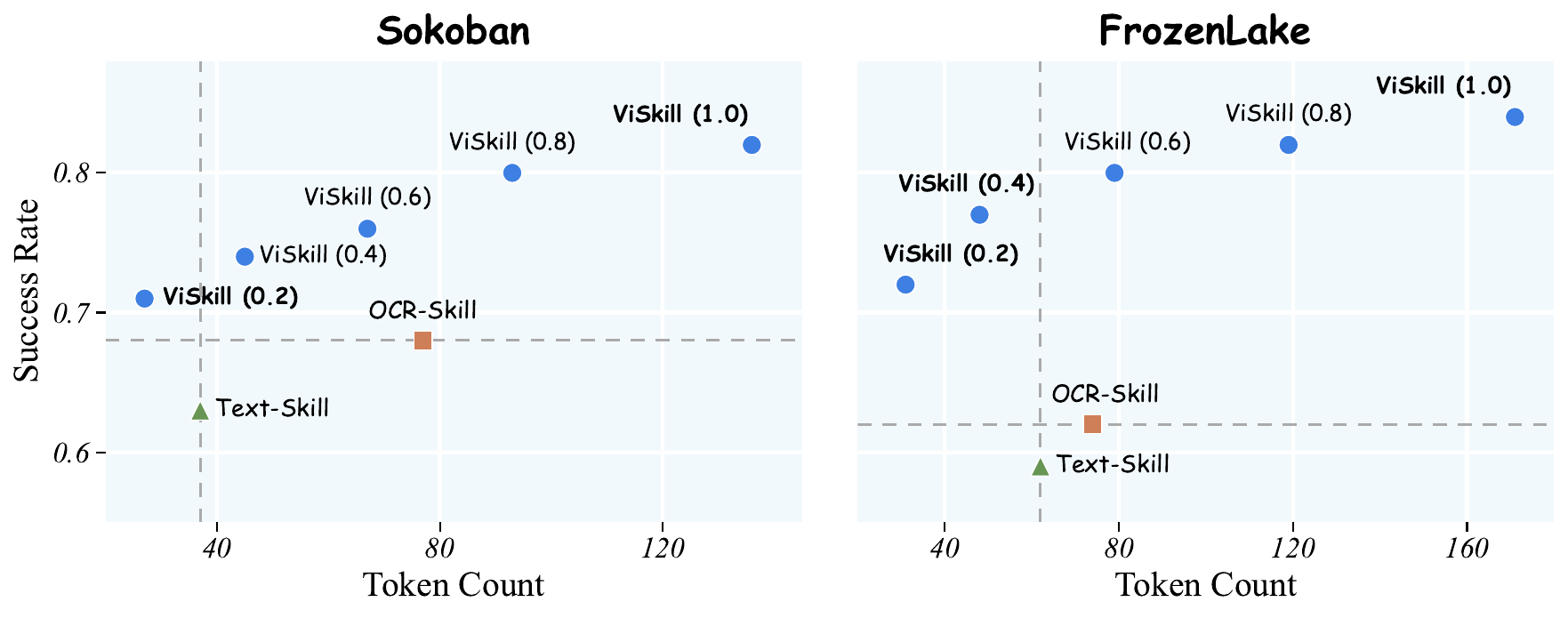}
        \vspace{-10pt}
        \caption{Success rate and token efficiency of visual skills at different card resolutions on Sokoban and FrozenLake, compared against the Text-Skill and OCR-Skill baselines.}
        \label{fig:visual_skill_efficiency}
    \end{minipage}
\end{figure*}

\paragraph{Cold-Start Budget.}
Figure~\ref{fig:cold_start_budget} investigates how the size of the cold-start library affects final performance. The cold-start ratio $\gamma$ scales the library size relative to $N_{\mathrm{conv}}$, the converged library size from a run without cold-start initialization. As $\gamma$ increases, performance improves consistently on both environments, indicating that a richer initial skill library accelerates and strengthens the feedback loop between skill distillation and policy learning. On Sokoban, performance saturates at $\gamma=0.6$ with a success rate of $1.00$, while FrozenLake continues to improve, reaching $0.89$ at $\gamma=1.0$. These results suggest that investing in a higher-quality cold-start library yields reliable gains, particularly in environments with more diverse task configurations.

\paragraph{Visual Skill Efficiency.}
Figure~\ref{fig:visual_skill_efficiency} examines the trade-off between skill card resolution and token efficiency. The scaling factor is applied to both the height and width of the skill card image, producing lower-resolution representations that consume fewer tokens. Across both environments, success rate increases monotonically with scale, with ViSkill~($1.0$) achieving the highest performance. Notably, even the most compact variant, ViSkill~($0.2$), surpasses both Text-Skill and OCR-Skill on Sokoban while consuming the fewest tokens, and on FrozenLake, ViSkill~($0.2$) and ViSkill~($0.4$) similarly outperform both baselines with substantially lower token consumption. This demonstrates that visual skill representations remain effective even at significantly reduced resolutions, offering a favorable efficiency-performance trade-off over text-based alternatives.

\paragraph{Evolved Skill Library Transfer.}
\begin{wrapfigure}{r}{0.48\columnwidth}
    \centering
    \includegraphics[width=\linewidth]{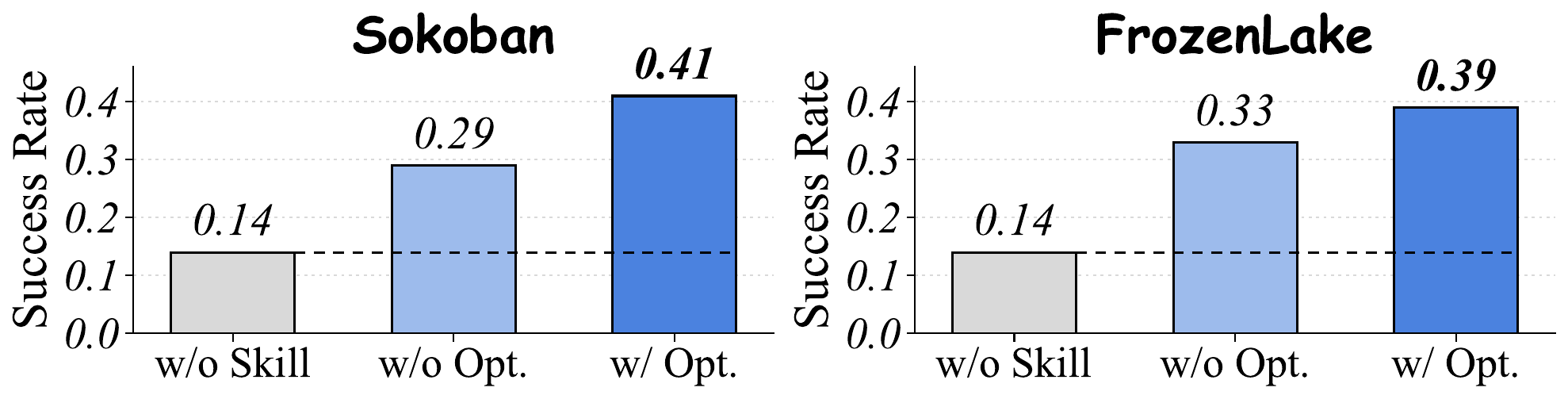}
    \caption{Transfer performance of evolved skill libraries on the unoptimized base model, with and without joint policy optimization.}
    \label{fig:evolved_skill_transfer}
\end{wrapfigure}
Figure~\ref{fig:evolved_skill_transfer} examines whether the skill library accumulated during training can transfer to the unoptimized base model. We compare two libraries: one distilled with joint policy optimization and one without. Augmenting the base model with either library yields consistent gains over the no-skill baseline, confirming that evolved skills are reusable beyond the trained policy. Skills accumulated with joint policy optimization further improve success rate to $0.41$ on Sokoban and $0.39$ on FrozenLake, outperforming those distilled without optimization. This demonstrates that the co-evolution of policy and skill library produces higher-quality skills: an improving policy generates more successful and diverse trajectories and strategies, which in turn enrich the library with more effective and transferable skills.

\section{Conclusion}

We present ViSkill, a visual-native skill learning framework that represents reusable procedural experience as composite visual skill cards and couples skill retrieval with reinforcement optimization through a closed feedback loop. By preserving spatial and geometric structure in a form directly accessible to VLM agents, ViSkill addresses the representation gap of text-centric skill approaches, while skill-guided rewards and online distillation enable policy improvement and skill accumulation to co-evolve. Experiments on Sokoban, FrozenLake, and PrimitiveSkill demonstrate consistent gains over proprietary and open-source baselines, and ablations confirm that visual representations, strategic guidance, and joint optimization each contribute to the positive feedback loop.

\paragraph{Limitations.} ViSkill is currently evaluated on environments supported by the VAGEN framework; whether visual skill cards provide consistent benefits in more complex or open-ended settings remains to be validated. The skill-guided reward further assumes that useful partial progress reaches states represented by the retrieved skill, which may provide limited guidance when valid solutions traverse disjoint state paths.

\subsection*{AI Use Statement}
We used generative AI tools to assist with implementation, figure plotting, and manuscript drafting and editing. All AI-assisted content was reviewed and verified by the authors, who take full responsibility for the final work.
\subsection*{Ethics Statement}
This work is conducted exclusively in simulated environments and involves no human participants, personal data, or physical systems. Real-world deployment would require task-specific safety constraints and human oversight. Trajectory-quality and novelty gates mitigate the accumulation of low-quality behaviors, though they do not replace safety validation in open-ended applications.

\subsection*{Reproducibility Statement}
We have made substantial efforts to ensure reproducibility. Appendix~\ref{sec:appendix_methodology_details} details the geometric descriptors, similarity functions, trajectory-length budgets, skill rendering, library management, and reward and PPO hyperparameters. Appendix~\ref{sec:appendix_training_details} reports environment configurations, training and evaluation seed ranges, and training hyperparameters. Appendix~\ref{sec:appendix_prompt_templates} provides the complete prompt templates for strategy distillation and environment interaction.

\bibliography{iclr2027_conference}
\bibliographystyle{iclr2027_conference}

\newpage

\appendix

\section{Methodology Details}
\label{sec:appendix_methodology_details}

\paragraph{Geometric Features and Similarity.}
Language-level task descriptions alone do not reliably identify a transferable skill in spatial environments: identical goals may require different action sequences under different layouts. We therefore retrieve skills using geometric descriptors derived from the initial configuration, capturing directional relations, distances, and local obstacle structure between task-relevant entities.

For discrete environments (Sokoban and FrozenLake), we partition the descriptor into semantically meaningful blocks and compute similarity as a weighted average of their exact-match rates:
\begin{equation}
    \operatorname{sim}(g, g_k) = \sum_{c \in \mathcal{C}} w_c \left( \frac{1}{|g^{(c)}|} \sum_{r=1}^{|g^{(c)}|} \mathbb{I}\left[g^{(c)}_r = g^{(c)}_{k,r}\right] \right),
\end{equation}
where $\mathcal{C}$ denotes the set of descriptor blocks and $\sum_{c \in \mathcal{C}} w_c = 1$. For Sokoban, descriptors encode the player--box and box--target spatial relations together with local wall indicators; for FrozenLake, they encode the start--goal relation and hole indicators along directions connecting the start and goal.

\begin{figure*}[t]
    \centering
    \includegraphics[width=0.95\textwidth]{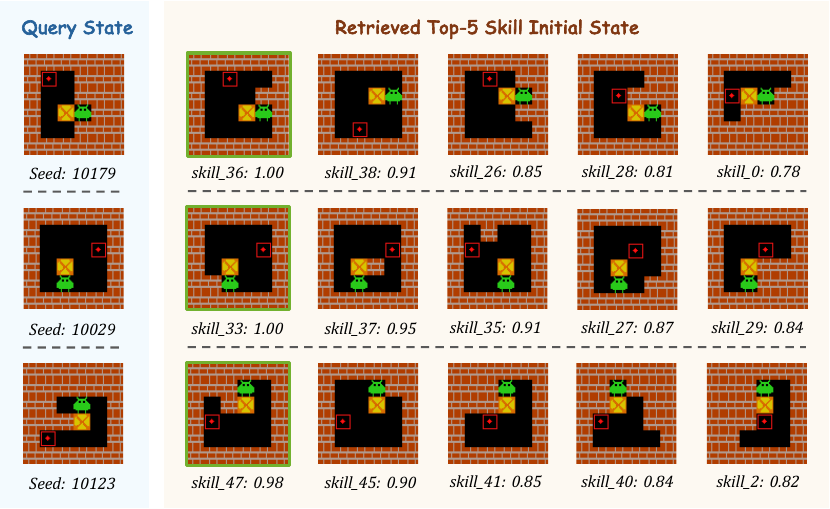}
    \caption{Geometry-based skill retrieval on three held-out Sokoban configurations. Each row shows a query state alongside its top-five retrieved skills, ordered by composite score, with the selected skill highlighted in green.}
    \label{fig:geometric_retrieval}
\end{figure*}

Figure~\ref{fig:geometric_retrieval} visualizes retrieval behavior on three held-out Sokoban configurations. In each case, the top-ranked skill exhibits the closest spatial arrangement, while lower-ranked candidates share progressively fewer task-relevant geometric relations.

For PrimitiveSkill, descriptors contain the $xy$ coordinates of task-relevant objects and a task identifier; skills are never retrieved across distinct manipulation tasks. Within the same task, we use a continuous similarity:

\begin{equation}
    \operatorname{sim}(g, g_k) = \max\!\left(0,\,1-\frac{\max_j\lVert p_j - p_{k,j}\rVert_2}{d_{\max}}\right),
\end{equation}

where $p_j$ and $p_{k,j}$ are the $xy$ coordinates of corresponding objects in the current and retrieved configurations, and $d_{\max}$ is a normalization constant.

\paragraph{Trajectory-Length Budgets and Solvers.}
The trajectory-length budget prevents successful but unnecessarily long trajectories from entering the skill library. In general, the quality gate uses a task-dependent budget $L_{\mathrm{bud}}(x)$; in our experiments we instantiate it with the shortest available solution length, $L_{\mathrm{bud}}(x) = L^{\star}(x)$, so a trajectory is admitted only if its length does not exceed the corresponding reference.

For Sokoban and FrozenLake, we obtain $L^{\star}(x)$ via breadth-first search over the grid state space, treating holes as impassable cells for FrozenLake. For PrimitiveSkill, each manipulation task has a fixed reference length given by its scripted oracle plan: $4$ actions for Place, Stack, and Align, $3$ for Drawer, and $6$ for Swap. The solvers are used solely to provide trajectory-quality budgets and optional cold-start demonstrations; they are not provided to the policy or used during skill retrieval.

\begin{figure*}[t]
    \centering
    \includegraphics[width=\textwidth]{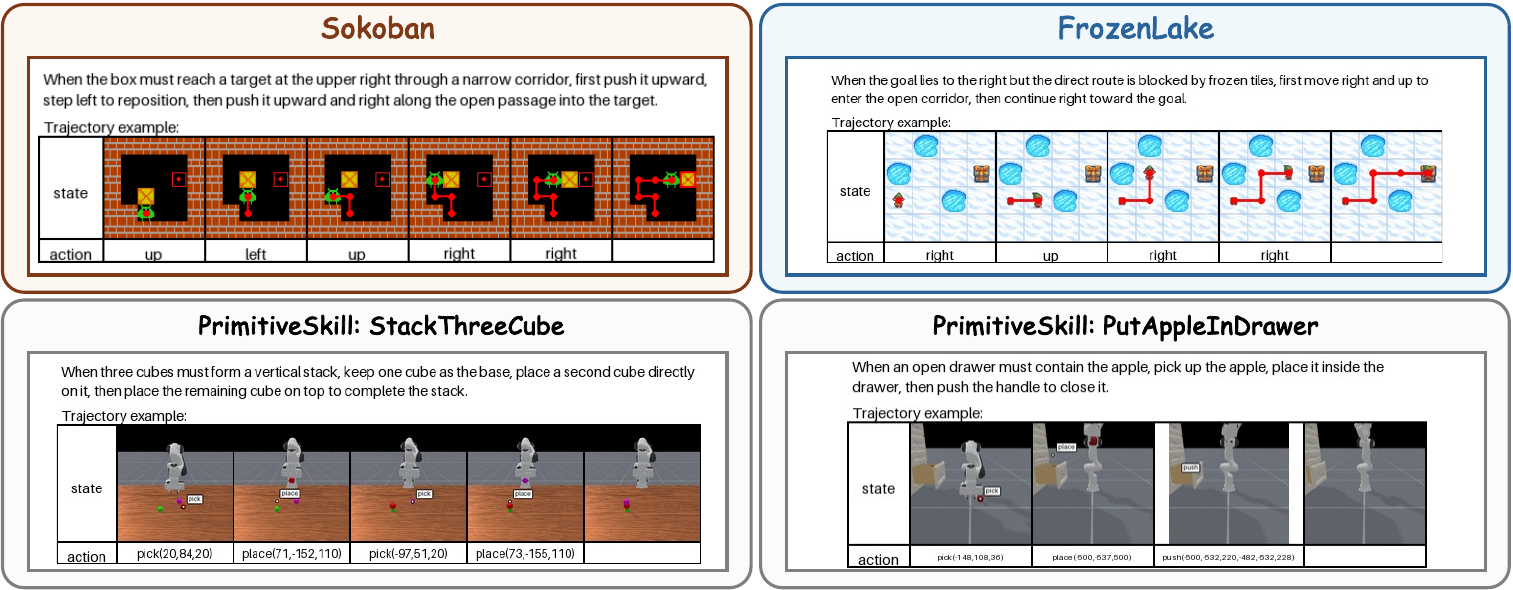}
    \caption{Representative visual skill cards from Sokoban, FrozenLake, and PrimitiveSkill. Each card combines a rendered strategy description with annotated trajectory frames and action labels into a single composite image.}
    \label{fig:skill_card_examples}
\end{figure*}

\paragraph{Skill Rendering.}
For grid environments, the renderer arranges trajectory frames in temporal order, augmenting them with aligned action labels and cumulative path overlays to clarify spatial progression. For PrimitiveSkill, the original third-person frames are preserved to retain object configurations and manipulation context. Figure~\ref{fig:skill_card_examples} shows representative visual skill cards across the three environments.

\paragraph{Skill Library Management.}
Each library entry stores the composite card, geometric descriptor, reference action trajectory, strategy description, utility, and usage count. The reference trajectory is retained for computing the skill-guidance reward; the composite card is appended to the VLM's visual context at decision time. Newly distilled skills are initialized with utility $0.5$ and usage count $0$; cold-start skills use the same initialization and are subsequently managed identically to online-distilled skills.

Only the skill retrieved for an episode receives a utility and usage-count update after termination; all non-retrieved skills retain their current statistics. When the library reaches capacity, we apply the utility--frequency eviction rule before admitting the new skill, so that infrequently used or ineffective skills are gradually displaced by more recent acquisitions.

\paragraph{Reward and Optimization.}
The environment reward is accumulated over interaction turns before the skill-guidance term is added. In Sokoban, each turn with a non-empty valid action sequence receives a formatting reward of $0.1$, and task completion receives a terminal reward of $1.0$. FrozenLake uses the same terminal reward with a formatting reward of $0.02$. In PrimitiveSkill, each valid turn receives a formatting reward of $0.1$, each newly completed manipulation stage (e.g., grasping, placing, or aligning an object) provides an incremental reward of $2.0$, and full task completion adds a terminal reward of $1.0$.

We instantiate $\ell$ using task-specific state progress rather than exact action alignment. In Sokoban and FrozenLake, progress is measured by relative position with respect to the initial state: the player--box configuration for Sokoban and the agent position for FrozenLake. In PrimitiveSkill, progress is determined by matching completed actions against the reference trajectory. In each case, $\ell$ is the furthest matching progress point along the retrieved skill trajectory.

For PPO optimization, the complete multi-turn interaction is tokenized into $\bar{\tau}$. Observation tokens, including image tokens from the environment and retrieved skill card, are excluded from both actor and critic losses; only policy-generated response tokens have $m_i=1$. We estimate advantages and returns via generalized advantage estimation with both its discount and smoothing parameters set to $1.0$. The actor is optimized with a clipped objective and $\epsilon=0.2$. The critic $V_\phi$ minimizes the clipped value loss:
\begin{equation}
    \begin{aligned}
        \mathcal{L}_{\mathrm{V}}(\phi)
        &=
        \frac{1}{2\sum_{i=1}^{N}m_i}
        \sum_{i=1}^{N}m_i
        \max\!\Bigg(
            \left(V_\phi(\bar{\tau}_{\leq i})-\widehat{G}_i\right)^2, \\
        &\qquad\quad
            \left(
                \operatorname{clip}\!\Bigl(
                    V_\phi(\bar{\tau}_{\leq i}),
                    V_{\mathrm{old},i}-\epsilon_{\mathrm{V}},
                    V_{\mathrm{old},i}+\epsilon_{\mathrm{V}}
                \Bigr)-\widehat{G}_i
            \right)^2
        \Bigg),
    \end{aligned}
\end{equation}
where $\widehat{G}_i$ is the GAE return, $V_{\mathrm{old},i}$ is the rollout value prediction, and $\epsilon_{\mathrm{V}}=0.5$. We use one PPO epoch per rollout batch with no entropy or KL auxiliary loss.

\section{Training and Evaluation Details}
\label{sec:appendix_training_details}

\subsection{Environment and Task Configurations}

All tasks provide the policy with RGB observations together with a textual task instruction and action-format specification; no symbolic state is exposed to the policy. Each episode is generated deterministically from its random seed, and success is evaluated by the task-specific environment predicate. Training and evaluation use disjoint seed ranges across all environments, ensuring that evaluation layouts and object configurations are never seen during training.

\paragraph{Sokoban.}
At each turn, the policy outputs up to three actions from \{\texttt{up}, \texttt{down}, \texttt{left}, \texttt{right}\}. An episode lasts at most five turns and succeeds when the box reaches its target.

\paragraph{FrozenLake.}
We use randomly generated maps with frozen-cell probability $0.8$. Transitions are deterministic, with slippery dynamics disabled. The policy selects from \{\texttt{up}, \texttt{down}, \texttt{left}, \texttt{right}\} and may output up to five actions per turn over at most five turns. An episode terminates when the agent reaches the goal or falls into a hole.

\paragraph{PrimitiveSkill.}
We use third-person RGB observations from ManiSkill~\citep{mu2021maniskill} across five tasks: Place, Stack, Drawer, Align, and Swap. The policy controls a Franka Panda arm via parameterized actions from \{\texttt{pick}$(x,y,z)$, \texttt{place}$(x,y,z)$, \texttt{push}$(x_1,y_1,z_1,x_2,y_2,z_2)$\} with continuous $xyz$ coordinates. The policy may issue up to two actions per turn, with a maximum of three turns for Place, Stack, Drawer, and Align, and four for Swap.

\subsection{Training Hyperparameters}

Our implementation builds on VAGEN~\citep{wang2026vagen}, which is developed on top of the veRL framework~\citep{sheng2025verl}. We share a common optimization and rollout configuration across all three environments, adapting only the training horizon, computational resources, and sequence-length limits to each task. Tables~\ref{tab:shared_training_hyperparameters} and~\ref{tab:environment_training_hyperparameters} summarize the shared and environment-specific hyperparameters, respectively.

\begin{table}[H]
    \centering
    \small
    \setlength{\tabcolsep}{6pt}
    \resizebox{0.5\linewidth}{!}{%
    \begin{tabular}{@{}lr@{}}
        \toprule
        \textbf{Hyperparameter} & \textbf{Value} \\
        \midrule
        \rowcolor{gray!10}
        \multicolumn{2}{@{}l@{}}{\textbf{\textit{Optimization}}} \\
        Policy backbone & Qwen2.5-VL-3B-Instruct \\
        Algorithm & PPO \\
        Advantage estimator & GAE \\
        PPO mini-batch size & $32$ \\
        Actor learning rate & $1\times10^{-6}$ \\
        Critic learning rate & $1\times10^{-5}$ \\
        Weight decay & $0.01$ \\
        GAE discount parameter & $1.0$ \\
        GAE smoothing parameter & $1.0$ \\
        PPO epochs per rollout batch & $1$ \\
        Actor clipping coefficient & $0.2$ \\
        Critic clipping coefficient & $0.5$ \\
        Entropy coefficient & $0.0$ \\
        KL coefficient & $0.0$ \\
        \midrule
        \rowcolor{gray!10}
        \multicolumn{2}{@{}l@{}}{\textbf{\textit{Rollout}}} \\
        Rollout batch size & $128$ \\
        Rollouts per task & $1$ \\
        GPU memory utilization & $0.6$ \\
        \midrule
        \rowcolor{gray!10}
        \multicolumn{2}{@{}l@{}}{\textbf{\textit{Skill System}}} \\
        Initial skill utility & $0.5$ \\
        Similarity weight $\alpha$ & $0.9$ \\
        EMA coefficient $\beta$ & $0.1$ \\
        Retrieval threshold $\delta_r$ & $0.8$ \\
        Novelty threshold $\delta_d$ & $0.9$ \\
        Skill-guidance weight $\lambda$ & $0.8$ \\
        \bottomrule
    \end{tabular}%
    }
    \caption{Shared training hyperparameters across all environments.}
    \label{tab:shared_training_hyperparameters}
\end{table}

\begin{table}[H]
    \centering
    \small
    \setlength{\tabcolsep}{5pt}
    \resizebox{0.62\linewidth}{!}{%
    \begin{tabular}{@{}lccc@{}}
        \toprule
        \textbf{Hyperparameter} & \textbf{Sokoban} & \textbf{FrozenLake} & \textbf{PrimitiveSkill} \\
        \midrule
        Total training steps & $300$ & $300$ & $60$ \\
        Training GPUs & $4$ & $2$ & $2$ \\
        Environment GPUs & -- & -- & $1$ \\
        Maximum prompt length & $5{,}000$ & $5{,}000$ & $3{,}000$ \\
        Maximum response length & $4{,}000$ & $4{,}000$ & $10{,}000$ \\
        \bottomrule
    \end{tabular}%
    }
    \caption{Environment-specific training hyperparameters.}
    \label{tab:environment_training_hyperparameters}
\end{table}

\section{Prompt Templates}
\label{sec:appendix_prompt_templates}

We provide the full prompt templates used for environment interaction and strategy distillation. Braced fields denote runtime-supplied values.

\subsection{Strategy Distillation Prompt}

This prompt generates the strategy description for a newly accepted skill, conditioned on the task description, initial observation, rendered trajectory, and environment-specific examples. During online self-distillation, we append the \texttt{<strategy>} prefix to elicit an extractable continuation, with generation capped at $256$ tokens. Cold-start distillation uses the same prompt body without the forced prefix and allows at most $512$ tokens from the external model.

\begin{AIbox}[breakable]{Strategy Distillation Prompt Template}
\setlength{\parindent}{0pt}\setlength{\parskip}{0pt}
[Skill Distillation]\par
Task: \{task\_description\}\par
\smallskip
You successfully completed this task. Based on the initial state and your trajectory, distill a reusable strategy that can guide solving similar tasks.\par
\smallskip
The strategy should describe WHEN to apply it (what the state looks like) and WHAT to do (concrete actions or approach).\par
\smallskip
Your response should be in the format of:\par
\texttt{<strategy>}...\texttt{</strategy>}\par
\smallskip
\{examples\}\par
Initial state:\par
\texttt{<image>}\par
Successful trajectory:\par
\texttt{<image>}\par
\smallskip
Now distill your strategy:\texttt{<strategy>}\par
\end{AIbox}

\subsection{Environment Interaction Prompts}

\paragraph{Sokoban Prompts.}
The following templates specify the Sokoban system prompt and initial instruction prompt. The retrieved composite skill card is inserted between the system prompt and the initial observation.

\begin{AIbox}[breakable]{Sokoban System Prompt Template}
\setlength{\parindent}{0pt}\setlength{\parskip}{0pt}
You are a Sokoban solver.\par
Sokoban Quick Guide\par
Goal: Push all boxes onto targets.\par
Symbols (If image is provided there are no symbols):\par
\# Wall | \_ Floor | O Target | X Box | P You | $\surd$ Box on Target | S You on Target\par
Rules:\par
1. Push boxes (can't pull).\par
2. Avoid walls.\par
Actions you can take: Left, Down, Right, Up.\par
You can take up to 3 action(s) at a time, separated by ,.\par
You should first give your reasoning, and then your answer.\par
Your response should be in the format of:\par
\texttt{<think>}...\texttt{</think>}\texttt{<answer>}...\texttt{</answer>}\par
\smallskip
Example 1:\par
\texttt{<think>}The box is one step below me, and the target is two steps below me. I should go down to reach the box and then push it down to the target.\texttt{</think>}\par
\texttt{<answer>}Down\texttt{</answer>}\par
\smallskip
Example 2:\par
\texttt{<think>}The box is to the right of me, and the target is further to the right. I need to move right to get behind the box and push it toward the target.\texttt{</think>}\par
\texttt{<answer>}Right\texttt{</answer>}\par
\smallskip
Example 3:\par
\texttt{<think>}The box is above me, and the target is above the box. I should move up to reach the box and then push it upward to the target.\texttt{</think>}\par
\texttt{<answer>}Up\texttt{</answer>}\par
\end{AIbox}

\begin{AIbox}[breakable]{Sokoban Instruction Prompt Template}
\setlength{\parindent}{0pt}\setlength{\parskip}{0pt}
[SYSTEM]\par
\{Sokoban system prompt above\}\par
\smallskip
[Skill Reference]:\par
The following shows a successful solution for a similar task. Use it as guidance for your approach.\par
\texttt{<image>}\par
\smallskip
[USER]\par
[Initial Observation]:\par
\texttt{<image>}\par
Decide your next action(s).\par
\end{AIbox}

\paragraph{FrozenLake Prompts.}
The following templates specify the FrozenLake system prompt and initial instruction prompt. The retrieved composite skill card is inserted between the system prompt and the initial observation.

\begin{AIbox}[breakable]{FrozenLake System Prompt Template}
\setlength{\parindent}{0pt}\setlength{\parskip}{0pt}
You are a FrozenLake solver.\par
FrozenLake Quick Guide\par
Goal: Reach the goal (G).\par
Symbols (If image is provided there are no symbols):\par
\_ Frozen | O Hole | G Goal | P Player | X Player fell into hole | V Player on goal\par
Rules:\par
1. Avoid falling into holes.\par
2. Frozen tiles are slippery, you may move perpendicular to your intended direction.\par
Actions you can take: Left, Down, Right, Up.\par
You can take up to 5 action(s) at a time, separated by ,.\par
You should first give your reasoning, and then your answer.\par
Your response should be in the format of:\par
\texttt{<think>}...\texttt{</think>}\texttt{<answer>}...\texttt{</answer>}\par
\smallskip
Example 1:\par
\texttt{<think>}The goal is below me. I should go down to reach it while avoiding the hole on my left.\texttt{</think>}\par
\texttt{<answer>}Down\texttt{</answer>}\par
\smallskip
Example 2:\par
\texttt{<think>}The goal is to my right and there's a hole directly below me. I should go right first to avoid the hole.\texttt{</think>}\par
\texttt{<answer>}Right\texttt{</answer>}\par
\smallskip
Example 3:\par
\texttt{<think>}I can see the goal is up and to the left. I should move up first to get closer.\texttt{</think>}\par
\texttt{<answer>}Up\texttt{</answer>}\par
\end{AIbox}

\begin{AIbox}[breakable]{FrozenLake Instruction Prompt Template}
\setlength{\parindent}{0pt}\setlength{\parskip}{0pt}
[SYSTEM]\par
\{FrozenLake system prompt above\}\par
\smallskip
[Skill Reference]:\par
The following shows a successful solution for a similar task. Use it as guidance for your approach.\par
\texttt{<image>}\par
\smallskip
[USER]\par
[Initial Observation]:\par
\texttt{<image>}\par
Decide your next action(s).\par
\end{AIbox}

\paragraph{PrimitiveSkill Prompts.}
The following templates specify the PrimitiveSkill system prompt and initial instruction prompt. The retrieved composite skill card is inserted between the system prompt and the initial observation.

\begin{AIbox}[breakable]{PrimitiveSkill System Prompt Template}
\setlength{\parindent}{0pt}\setlength{\parskip}{0pt}
You are an AI assistant controlling a Franka Emika robot arm. Your goal is to understand human instructions and translate them into a sequence of executable actions for the robot, based on visual input and the instruction.\par
\smallskip
Action Space Guide\par
You can command the robot using the following actions:\par
\smallskip
1. pick(x, y, z) \# To grasp an object located at position(x,y,z) in the robot's workspace.\par
2. place(x, y, z) \# To place the object currently held by the robot's gripper at the target position (x,y,z).\par
3. push(x1, y1, z1, x2, y2, z2) \# To push an object from position (x1,y1,z1) to (x2,y2,z2).\par
\smallskip
Hints:\par
1. The coordinates (x, y, z) are in millimeters and are all integers.\par
2. Please ensure that the coordinates are within the workspace limits.\par
3. The position is the center of the object, when you place, please consider the volume of the object. It's always fine to set z much higher when placing an item.\par
4. We will provide the object positions to you, but you need to match them to the object in the image by yourself. You're facing toward the negative x-axis, and the negative y-axis is to your left, the positive y-axis is to your right, and the positive z-axis is up.\par
\smallskip
Examples:\par
round1:\par
image1\par
Human Instruction: Put red cube on green cube and yellow cube on left target\par
Object positions:\par
[(62,-55,20),(75,33,20),(-44,100,20),(100,-43,0),(100,43,0)]\par
Reasoning: I can see from the picture that the red cube is on my left and green cube is on my right and near me.\par
Since I'm looking toward the negative x axis, and negative y-axis is to my left, (62,-55,20) would be the position of the red cube, (75,33,20) would be the position of the green cube and (-44,100,20) is the position of the yellow cube.\par
Also the (100,-43,0) would be the position of the left target, and (100,43,0) would be the position of the right target.\par
I need to pick up red cube first and place it on the green cube, when placing, I should set z much higher.\par
Answer: pick(62,-55,20)|place(75,33,50)\par
round2:\par
image2\par
Human Instruction: Put red cube on green cube and yellow cube on left target\par
Object positions:\par
[(75,33,50),(75,33,20),(-44,100,20),(100,-43,0),(100,43,0)]\par
Reasoning: Now the red cube is on the green cube, so I need to pick up the yellow cube and place it on the left target.\par
Answer: pick(-44,100,20)|place(100,-43,50)\par
\smallskip
You can take up to 2 action(s) at a time, separated by '|'.\par
You should first give your thought process, and then your answer.\par
Your response should be in the format of:\par
\texttt{<think>}...\texttt{</think>}\texttt{<answer>}...\texttt{</answer>}\par
\smallskip
Example:\par
\texttt{<think>}I need to pick the red cube at (62,-55,20) and place it on the green cube at (75,33,20). When placing, I should set z much higher to stack properly.\texttt{</think>}\par
\texttt{<answer>}pick(62,-55,20)|place(75,33,50)\texttt{</answer>}\par
\end{AIbox}

\begin{AIbox}[breakable]{PrimitiveSkill Instruction Prompt Template}
\setlength{\parindent}{0pt}\setlength{\parskip}{0pt}
[SYSTEM]\par
\{PrimitiveSkill system prompt above\}\par
\smallskip
[Skill Reference]:\par
The following shows a successful solution for a similar task. Use it as guidance for your approach.\par
\texttt{<image>}\par
\smallskip
[USER]\par
\smallskip
[Initial Observation]:\par
\texttt{<image>}\par
Human Instruction: \{instruction\}\par
x\_workspace\_limit: \{x\_workspace\}\par
y\_workspace\_limit: \{y\_workspace\}\par
z\_workspace\_limit: \{z\_workspace\}\par
Object positions:\par
\{object\_positions\}\par
Other information:\par
\{other\_information\}\par
Decide your next action(s).\par
You can take up to 2 action(s) at a time, separated by '|'.\par
You should first give your thought process, and then your answer.\par
Your response should be in the format of:\par
\texttt{<think>}...\texttt{</think>}\texttt{<answer>}...\texttt{</answer>}\par
\smallskip
Example:\par
\texttt{<think>}I need to pick the red cube at (62,-55,20) and place it on the green cube at (75,33,20). When placing, I should set z much higher to stack properly.\texttt{</think>}\par
\texttt{<answer>}pick(62,-55,20)|place(75,33,50)\texttt{</answer>}\par
\end{AIbox}

\begin{figure*}[!h]
    \centering
    \includegraphics[width=\textwidth]{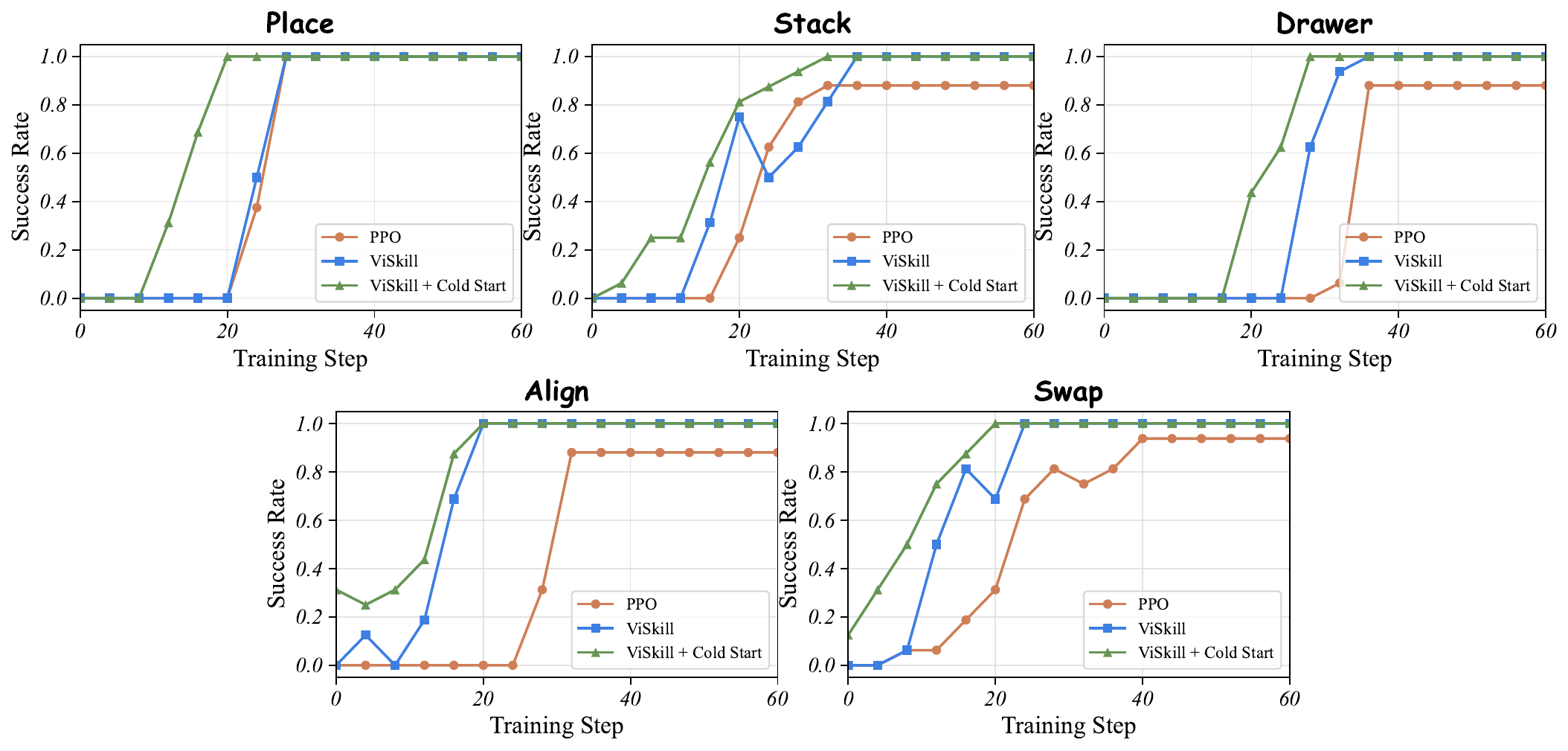}
    \caption{Task-wise success-rate dynamics across the five PrimitiveSkill manipulation tasks. Both \method variants reach full success faster than PPO, with cold-start initialization further accelerating early-stage learning.}
    \label{fig:primitive_task_dynamics}
\end{figure*}

\begin{figure*}[!h]
    \centering
    \includegraphics[width=0.76\textwidth]{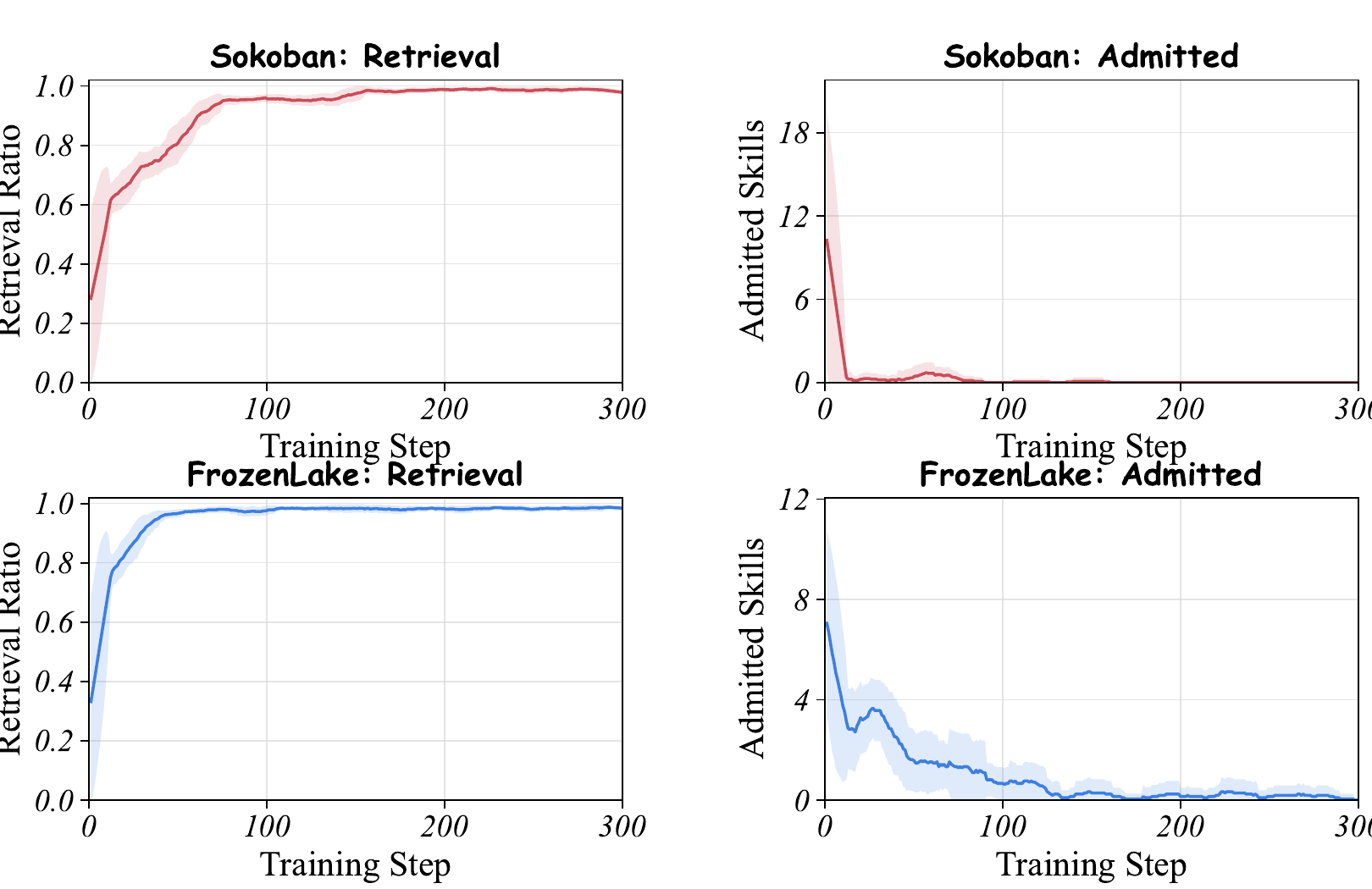}
    \caption{Skill retrieval and distillation dynamics on Sokoban and FrozenLake. Retrieval fraction (left) and newly admitted skills per step (right). Retrieval becomes nearly ubiquitous as the library grows, while admissions diminish once common configurations are covered.}
    \label{fig:skill_usage_dynamics}
\end{figure*}

\section{Additional Analysis}
\label{sec:appendix_additional_analysis}

\subsection{Additional Training Dynamics}

\paragraph{PrimitiveSkill Task-Wise Success-Rate Dynamics.}

Figure~\ref{fig:primitive_task_dynamics} disaggregates the average PrimitiveSkill result across its five manipulation tasks. \method consistently reaches full success earlier than PPO on all five tasks. PPO eventually solves Place but plateaus below full success on the remainder, whereas both \method variants achieve a success rate of $1.00$. Cold-Start further accelerates early-stage learning across tasks. These results confirm that visual skills provide reusable guidance across diverse coordinate-grounded manipulation objectives, rather than improving only the aggregate metric.

\paragraph{Skill Retrieval and Distillation Dynamics.}

Figure~\ref{fig:skill_usage_dynamics} illustrates the closed-loop evolution of the skill library. In both environments, the retrieval fraction rises rapidly and stabilizes near one as the library gains coverage, reflecting that existing skills increasingly satisfy the geometric retrieval criterion. Skill admissions, by contrast, concentrate in the early stage and taper off thereafter. This separation confirms that the novelty gate prevents redundant expansion: once common configurations are covered, the agent reuses existing skills while admitting only genuinely novel trajectories.

\begin{figure*}[t]
    \centering
    \includegraphics[width=\textwidth]{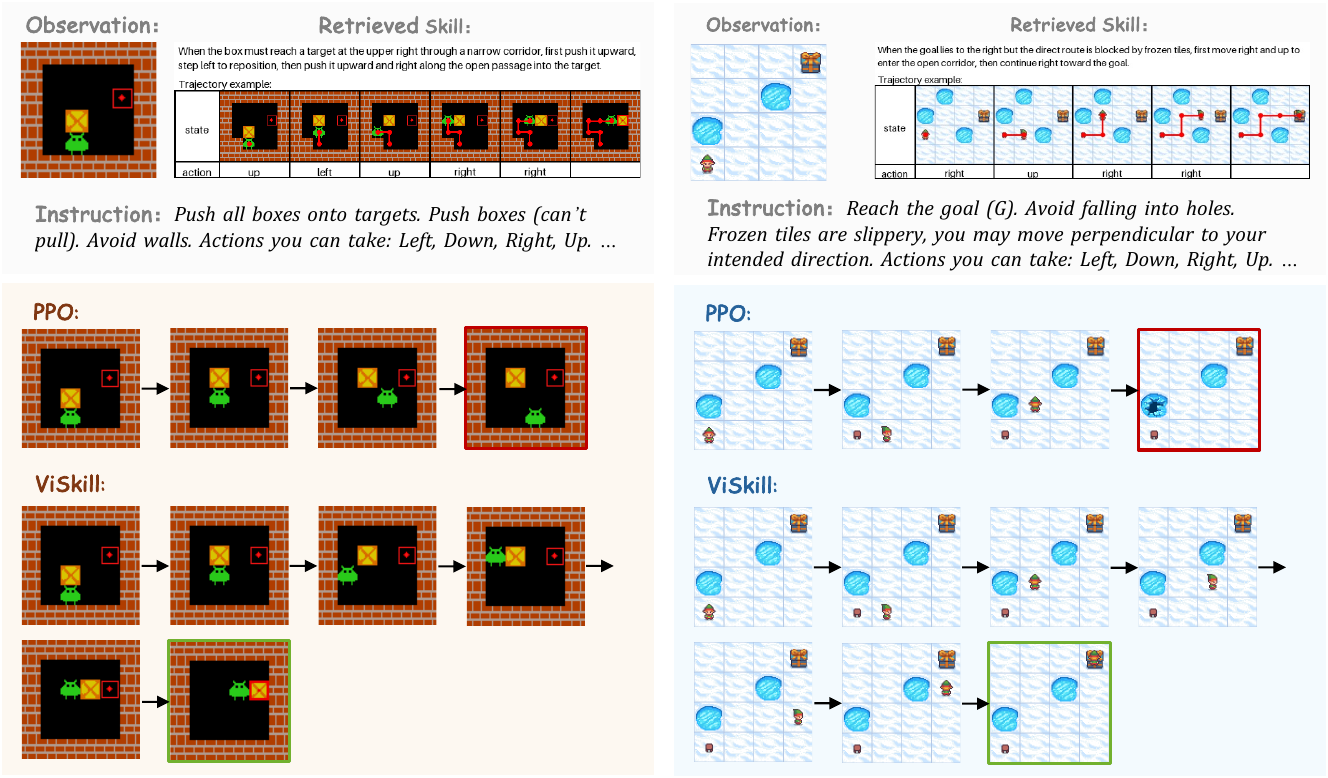}
    \vspace{-10pt}
    \caption{Qualitative comparison with PPO on Sokoban and FrozenLake. For each configuration, PPO reaches a failure state while \method retrieves a relevant visual skill card and completes the task.}
    \label{fig:case_study}
\end{figure*}
\vspace{-10pt}

\subsection{Case Study}

Figure~\ref{fig:case_study} compares PPO and \method on matched Sokoban and FrozenLake configurations. In both cases, the retrieved visual skill provides a spatially grounded reference that guides the policy toward a productive direction and encourages successful multi-turn behavior. PPO reaches failure states under the same configurations.

In the FrozenLake example, the trajectory stored in the retrieved skill does not exactly match the one executed by \method. The policy takes a different valid route while remaining consistent with the spatial structure and high-level intent of the skill card, indicating that \method uses the card as procedural guidance rather than replaying reference actions.

\subsection{Quality Threshold Ablation}
\label{sec:appendix_quality_threshold}

\begin{wraptable}{r}{0.38\columnwidth}
    \vspace{-12pt}
    \centering
    \small
    \setlength{\tabcolsep}{3pt}
    \begin{tabular}{lcc}
        \toprule
        \textbf{Threshold} & \textbf{Sokoban} & \textbf{FrozenLake} \\
        \midrule
        Solver-Based & 0.82 & 0.84 \\
        Static-10 & 0.84 & 0.78 \\
        \bottomrule
    \end{tabular}
    \caption{Success rates under different quality thresholds.}
    \vspace{-15pt}
    \label{tab:quality_threshold}
\end{wraptable}

Table~\ref{tab:quality_threshold} examines whether \method depends critically on solver-derived quality thresholds by substituting them with a fixed trajectory-length budget. Static-10 achieves success rates of $0.84$ on Sokoban and $0.78$ on FrozenLake, against $0.82$ and $0.84$ for solver-based filtering. The fixed threshold slightly outperforms the solver on Sokoban while incurring only a modest decrease on FrozenLake, indicating that solver access is not a prerequisite for effective quality control. A simple fixed budget therefore provides a practical alternative when task-specific solvers are unavailable.

\subsection{Feature Extractor Ablation}
\label{sec:appendix_feature_extractors}

\begin{wraptable}{r}{0.48\columnwidth}
    \vspace{-12pt}
    \centering
    \small
    \setlength{\tabcolsep}{3pt}
    \begin{tabular}{lcc}
        \toprule
        \textbf{Extractor} & \textbf{Sokoban} & \textbf{FrozenLake} \\
        \midrule
        Rule-Based & 0.82 & 0.84 \\
        DINOv2 & 0.80 & 0.83 \\
        \bottomrule
    \end{tabular}
    \caption{Success rates with different feature extractors.}
    \vspace{-12pt}
    \label{tab:feature_extractors}
\end{wraptable}

Table~\ref{tab:feature_extractors} investigates the role of hand-crafted geometric descriptors by replacing them with DINOv2~\citep{oquab2023dinov2} visual features for skill retrieval. Despite receiving no task-specific supervision, DINOv2 achieves success rates of $0.80$ on Sokoban and $0.83$ on FrozenLake, falling short of the rule-based extractor by only $0.02$ and $0.01$ in absolute terms. This narrow gap demonstrates that the spatial structure needed for reliable retrieval is largely recoverable from general-purpose visual representations, and that \method is not fundamentally tied to hand-crafted descriptors. Replacing them with learned encoders thus emerges not merely as a future direction but as a practically viable alternative, broadening the applicability of \method to environments where geometric features are difficult to specify manually.

\end{document}